\documentclass{article}
\def\HybridCUAAppendixPreamble{1}
\ifdefined\HybridCUAAppendixPreamble
    \PassOptionsToPackage{table}{xcolor}
    \RequirePackage{listings}
    \RequirePackage{tcolorbox}
    \tcbuselibrary{skins}
    \expandafter 
\fi

\section{Limitations}
\label{app:limitations}

HybridCUA has several limitations. First, its benefits depend on the
availability and stability of command-line interfaces. Applications without
usable CLIs, restricted-shell environments, or operating-system-specific
command semantics may require substantially different interface-routing
behavior. Although we evaluate transfer on OSWorld-MCP and
WindowsAgentArena, these benchmarks do not cover the full diversity of
applications, operating systems, permission settings, and long-horizon
workflows. Second, our data construction currently focuses primarily on the
applications included in OSWorld. Although these applications cover common
desktop workflows, they do not fully represent the diversity of
domain-specific applications or out-of-distribution environments encountered
in real-world use. Future work should extend the data construction pipeline
to a broader range of specialized applications and synthesize more diverse
OOD environments for HybridCUA training, thereby improving its generalization
to unseen applications, workflows, and interface configurations.

\section{Action Space and Interaction Protocol}
\label{app:action_space}

Table~\ref{tab:action_space} specifies the executable and control actions.
The shared \texttt{bash} interface is an action representation: using this
wrapper does not make a PyAutoGUI interaction a CLI operation. GUI actions
manipulate the visible interface through PyAutoGUI, whereas direct CLI
commands operate through the shell and the programs it invokes.

\begin{center}
    \refstepcounter{table}
    \label{tab:action_space}
    {\small \textbf{Table~\thetable:} Action space in HybridCUA.\par}
    \vspace{\abovecaptionskip}
    \scriptsize
    \setlength{\tabcolsep}{3pt}
    \renewcommand{\arraystretch}{1.05}
    \begin{tabular}{@{}p{0.12\linewidth}p{0.47\linewidth}p{0.35\linewidth}@{}}
        \hline
        Action & Implementation & Definition \\
        \hline
        \multicolumn{3}{@{}l}{\textit{Executable Action}} \\
        \texttt{bash}
        & \texttt{bash(command, timeout)}
        & Executes either a GUI or CLI command through a single shell
          interface. \\
        \quad GUI
        & \shortstack[l]{\texttt{python3 <<'PY'}\\
          \texttt{import pyautogui}\\
          \texttt{\ldots}\\
          \texttt{PY}}
        & Executes one or more \texttt{pyautogui} calls in a quoted Python
          heredoc. \\
        &
        \texttt{pyautogui.click(x, y)}
        & Left-clicks at $(x,y)$. \\
        &
        \texttt{pyautogui.doubleClick(x, y)}
        & Double-clicks at $(x,y)$. \\
        &
        \texttt{pyautogui.tripleClick(x, y)}
        & Triple-clicks at $(x,y)$. \\
        &
        \texttt{pyautogui.rightClick(x, y)}
        & Right-clicks at $(x,y)$. \\
        &
        \texttt{pyautogui.middleClick(x, y)}
        & Middle-clicks at $(x,y)$. \\
        &
        \texttt{pyautogui.moveTo(x, y)}
        & Moves the cursor to $(x,y)$. \\
        &
        \texttt{pyautogui.dragTo(x, y, duration=0.5)}
        & Drags from the current position to $(x,y)$. \\
        &
        \texttt{pyautogui.mouseDown()}
        & Presses and holds the mouse button. \\
        &
        \texttt{pyautogui.mouseUp()}
        & Releases the mouse button. \\
        &
        \texttt{pyautogui.scroll(-5)}
        & Scrolls downward; a positive value scrolls upward. \\
        &
        \texttt{pyautogui.press('enter')}
        & Presses a single key. \\
        &
        \texttt{pyautogui.hotkey('ctrl', 's')}
        & Presses a keyboard shortcut. \\
        &
        \texttt{pyautogui.typewrite('text', interval=0.02)}
        & Types the specified text. \\
        &
        \texttt{pyautogui.keyDown('shift')}
        & Presses and holds a modifier key. \\
        &
        \texttt{pyautogui.keyUp('shift')}
        & Releases a modifier key. \\
        \quad CLI
        & Direct shell command
        & Executes a command in the shell; its stdout/stderr is paired with
          the subsequent screenshot. \\
        \multicolumn{3}{@{}l}{\textit{Interaction and Control Actions}} \\
        \texttt{wait}
        & \texttt{wait(time)}
        & Waits for the interface to stabilize. \\
        \texttt{terminate}
        & \texttt{terminate(status)}
        & Ends the task with a status of success or failure. \\
        \texttt{answer}
        & \texttt{answer(text)}
        & Submits a textual answer and finishes the task. \\
        \hline
    \end{tabular}
\end{center}

\label{app:interaction_protocol}
At decision step $t$, the agent receives the current screenshot $I_t$ and,
when the preceding action is a direct CLI command, its stdout/stderr
$\tilde{y}_{t-1}$. The task instruction and preceding interaction history
provide the context for generating the next action. After execution, the
environment returns a post-action screenshot together with CLI output when
available. Thus, CLI only trajectories restrict the \emph{action interface},
not the observation modality: they can still contain visual observations.

The executable action \texttt{bash(command, timeout)} accepts either a direct
shell command or a quoted Python heredoc containing PyAutoGUI operations.
The control actions \texttt{wait}, \texttt{terminate}, and \texttt{answer}
respectively allow the interface to settle, end an episode with a status, or
submit a final textual response. A model-generated completion declaration is
distinct from the task verifier's assessment of the resulting state.

\section{Dataset Construction and Statistics}
\label{app:dataset_details}

HybridCUA-8K comprises two different data units: a corpus of 5{,}023
supervised trajectories and a pool of 3{,}000 verified RL tasks. A trajectory
records one execution, while an RL task specifies an environment and a
verifiable objective from which multiple rollouts can be sampled.

\subsection{Supervised Fine-Tuning Data Construction}
\label{app:sft_data}

The SFT corpus contains 5{,}023 trajectories spanning 11 application domains
and three complementary interaction modes: GUI only, CLI only, and hybrid
GUI--CLI execution. LibreOffice Impress, Writer, and Calc contribute 866,
818, and 811 trajectories, respectively, while the multi-application split
contributes 786. Trajectories contain 10.4 logical steps on average, with a
median of 7, a 90th percentile of 22, and a maximum of 50.
Figure~\ref{fig:sft_trajectory_distribution} summarizes the domain and
modality composition together with the trajectory-length distribution. We
construct each trajectory type as follows.

\paragraph{GUI only trajectories.}
The GUI only split is derived from UI-MOPD
trajectories~\citep{lian2026uimopd}. Figure~\ref{fig:gui_only_case} shows
selected steps from a representative trajectory that changes a presentation
slide background from the default color to red. To preserve the original GUI
supervision while matching the unified action space, each source operation, including
clicking, dragging, scrolling, keyboard shortcuts, and text entry, is
translated into an equivalent PyAutoGUI call. For coordinate-based operations,
the model performs visual grounding in a \(1000\times1000\) coordinate space
shared by its visual input and predicted GUI locations; the predicted
coordinates are then mapped to the environment display coordinates at
execution time. The converted calls are serialized as quoted Python heredocs
inside the same \texttt{bash(command, timeout)} wrapper used by the other
trajectory types. For example, a source click action becomes a
\texttt{pyautogui.click(x, y)} call inside this wrapper. The wrapper standardizes
action serialization and execution but does not change the interface category:
PyAutoGUI-based interface manipulation remains a GUI action, whereas only a
command that directly invokes a shell program or an application CLI is
classified as a CLI action.

\begin{figure}[!ht]
    \centering
    \includegraphics[width=0.94\linewidth]{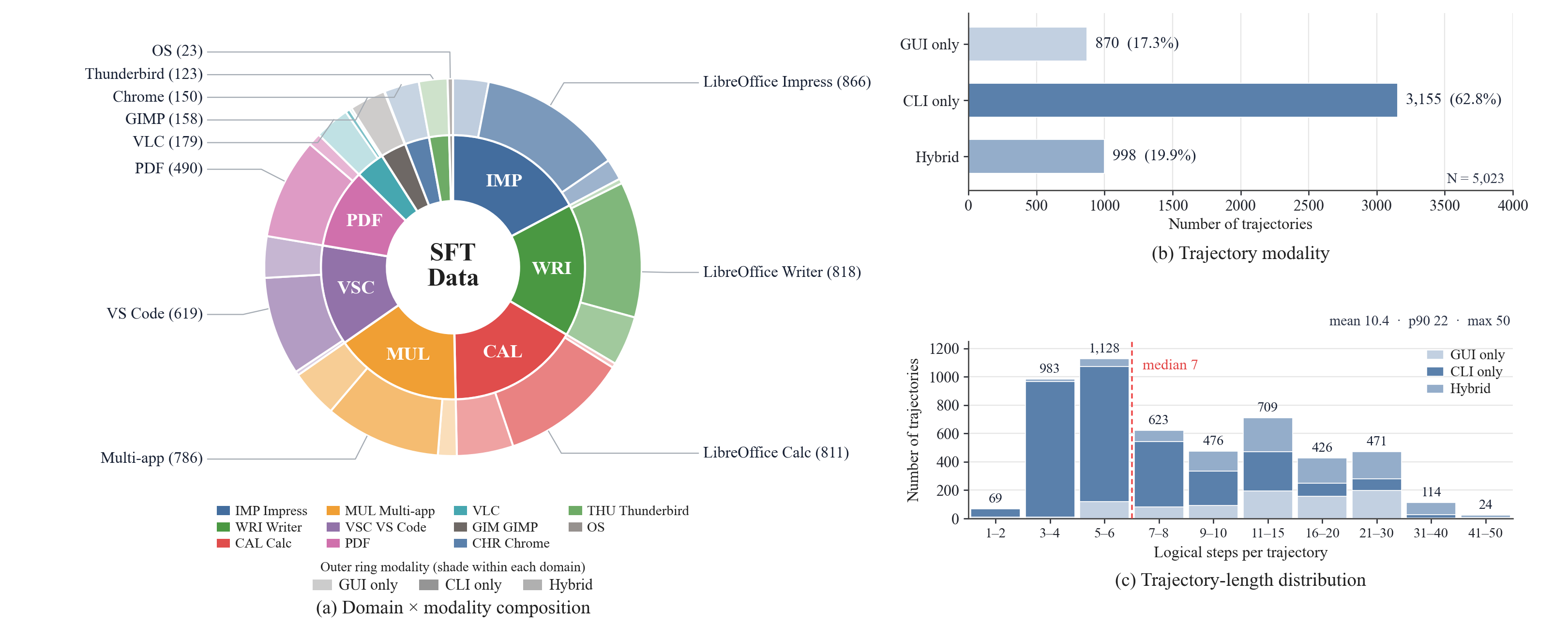}
    \caption{\textbf{Composition of the supervised fine-tuning corpus.}
    \textbf{(a)} Application-domain totals, with the outer-ring shade indicating
    GUI only, CLI only, or hybrid interaction within each domain.
    \textbf{(b)} Overall trajectory counts by modality.
    \textbf{(c)} Distribution of logical steps per trajectory, stacked by
    modality.}
    \label{fig:sft_trajectory_distribution}
\end{figure}

\par\medskip
\begingroup
\setlength{\parindent}{0pt}
\setlength{\parskip}{0pt}
\setlength{\fboxsep}{4pt}
\setlength{\fboxrule}{0.45pt}
\newsavebox{\guiCodeBox}
\newcommand{\guiConversionArrow}{%
    \colorbox{blue!7}{\textcolor{blue!65!black}{\Large$\boldsymbol{\Rightarrow}$}}%
}

\noindent
\begin{minipage}[c]{0.44\linewidth}
\centering
\begin{lrbox}{\guiCodeBox}
\begin{minipage}{0.92\linewidth}
\setlength{\topsep}{0pt}
\setlength{\partopsep}{0pt}
\color{black!78}\fontsize{5.8}{6.8}\selectfont
\begin{verbatim}
<tool_call>
<function=computer_use>
<parameter=action>
left_click
</parameter>
<parameter=coordinate>
[906, 287]
</parameter>
</function>
</tool_call>
\end{verbatim}
\end{minipage}
\end{lrbox}
\fcolorbox{black!18}{black!3}{\usebox{\guiCodeBox}}
\end{minipage}%
\hfill
\begin{minipage}[c]{0.08\linewidth}
\centering\guiConversionArrow
\end{minipage}%
\hfill
\begin{minipage}[c]{0.44\linewidth}
\centering
\begin{lrbox}{\guiCodeBox}
\begin{minipage}{0.92\linewidth}
\setlength{\topsep}{0pt}
\setlength{\partopsep}{0pt}
\color{blue!45!black}\fontsize{5.8}{6.8}\selectfont
\begin{verbatim}
<tool_call>
<function=computer_use>
<parameter=action>
bash
</parameter>
<parameter=command>
python3 <<'PY'
import pyautogui
pyautogui.click(906, 287)
PY
</parameter>
</function>
</tool_call>
\end{verbatim}
\end{minipage}
\end{lrbox}
\fcolorbox{blue!35!black}{blue!4}{\usebox{\guiCodeBox}}
\end{minipage}

\par\vspace{10pt}

\noindent
\begin{minipage}[c]{0.44\linewidth}
\centering
\begin{lrbox}{\guiCodeBox}
\begin{minipage}{0.92\linewidth}
\setlength{\topsep}{0pt}
\setlength{\partopsep}{0pt}
\color{black!78}\fontsize{5.8}{6.8}\selectfont
\begin{verbatim}
<tool_call>
<function=computer_use>
<parameter=action>
left_click_drag
</parameter>
<parameter=coordinate>
[593, 401]
</parameter>
<parameter=start_coordinate>
[456, 401]
</parameter>
</function>
</tool_call>
\end{verbatim}
\end{minipage}
\end{lrbox}
\fcolorbox{black!18}{black!3}{\usebox{\guiCodeBox}}
\end{minipage}%
\hfill
\begin{minipage}[c]{0.08\linewidth}
\centering\guiConversionArrow
\end{minipage}%
\hfill
\begin{minipage}[c]{0.44\linewidth}
\centering
\begin{lrbox}{\guiCodeBox}
\begin{minipage}{0.92\linewidth}
\setlength{\topsep}{0pt}
\setlength{\partopsep}{0pt}
\color{blue!45!black}\fontsize{5.8}{6.8}\selectfont
\begin{verbatim}
<tool_call>
<function=computer_use>
<parameter=action>
bash
</parameter>
<parameter=command>
python3 <<'PY'
import pyautogui
pyautogui.moveTo(456, 401)
pyautogui.dragTo(593, 401, duration=0.5)
PY
</parameter>
</function>
</tool_call>
\end{verbatim}
\end{minipage}
\end{lrbox}
\fcolorbox{blue!35!black}{blue!4}{\usebox{\guiCodeBox}}
\end{minipage}

\par\vspace{10pt}

\noindent
\begin{minipage}[c]{0.44\linewidth}
\centering
\begin{lrbox}{\guiCodeBox}
\begin{minipage}{0.92\linewidth}
\setlength{\topsep}{0pt}
\setlength{\partopsep}{0pt}
\color{black!78}\fontsize{5.8}{6.8}\selectfont
\begin{verbatim}
<tool_call>
<function=computer_use>
<parameter=action>
left_click
</parameter>
<parameter=coordinate>
[202, 223]
</parameter>
</function>
</tool_call>
<tool_call>
<function=computer_use>
<parameter=action>
key_down
</parameter>
<parameter=keys>
["shift"]
</parameter>
</function>
</tool_call>
<tool_call>
<function=computer_use>
<parameter=action>
left_click
</parameter>
<parameter=coordinate>
[202, 464]
</parameter>
</function>
</tool_call>
<tool_call>
<function=computer_use>
<parameter=action>
key_up
</parameter>
<parameter=keys>
["shift"]
</parameter>
</function>
</tool_call>
\end{verbatim}
\end{minipage}
\end{lrbox}
\fcolorbox{black!18}{black!3}{\usebox{\guiCodeBox}}
\end{minipage}%
\hfill
\begin{minipage}[c]{0.08\linewidth}
\centering\guiConversionArrow
\end{minipage}%
\hfill
\begin{minipage}[c]{0.44\linewidth}
\centering
\begin{lrbox}{\guiCodeBox}
\begin{minipage}{0.92\linewidth}
\setlength{\topsep}{0pt}
\setlength{\partopsep}{0pt}
\color{blue!45!black}\fontsize{5.8}{6.8}\selectfont
\begin{verbatim}
<tool_call>
<function=computer_use>
<parameter=action>
bash
</parameter>
<parameter=command>
python3 <<'PY'
import pyautogui
pyautogui.click(202, 223)
pyautogui.keyDown('shift')
pyautogui.click(202, 464)
pyautogui.keyUp('shift')
PY
</parameter>
</function>
</tool_call>
\end{verbatim}
\end{minipage}
\end{lrbox}
\fcolorbox{blue!35!black}{blue!4}{\usebox{\guiCodeBox}}
\end{minipage}

\endgroup

\par\medskip

\begin{center}
\begin{minipage}{\linewidth}
    \centering
    \includegraphics[width=\linewidth]{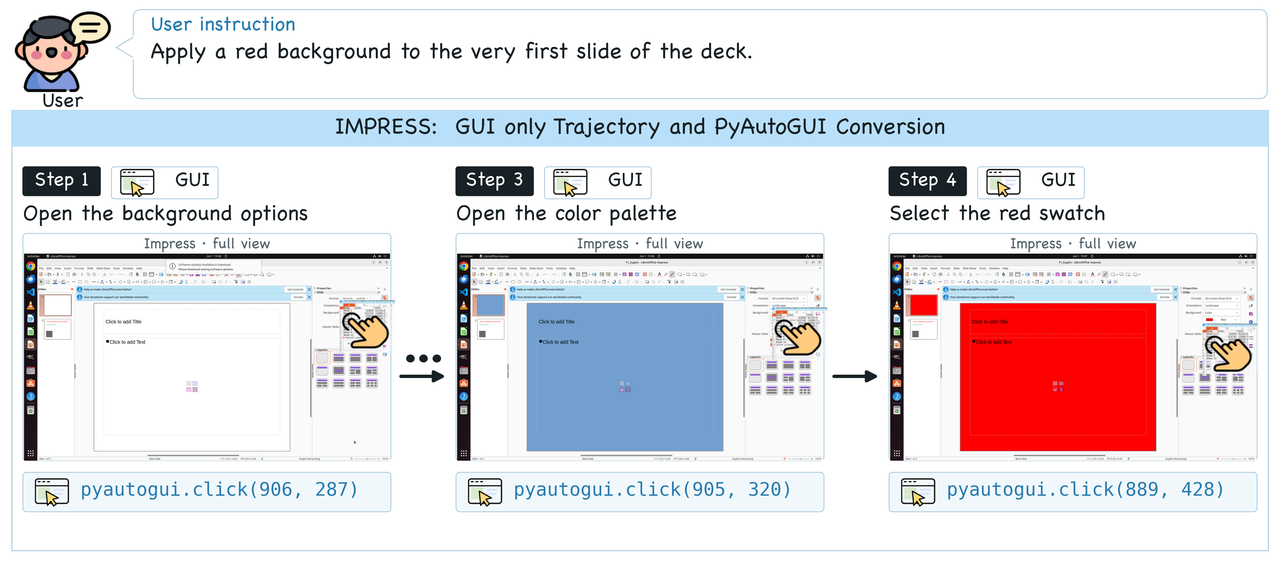}
    \refstepcounter{figure}\label{fig:gui_only_case}
    \vspace{\abovecaptionskip}

    {\small \textbf{Figure~\thefigure:} A representative GUI only trajectory
    paired with its direct PyAutoGUI representation.}
\end{minipage}
\end{center}

\paragraph{CLI only trajectories.}
We construct application-specific CLI skills from online application
documentation and reference implementations, and use
Qwen3.8-27B~\citep{qwen2026qwen38} equipped
with these skills and a Claude Code harness to sample trajectories in
CUA-Gym~\citep{wang2026cuagym}. Figure~\ref{fig:cli_only_case} shows selected steps from an
Impress example. Using Python/UNO, the agent identifies slides 1 and 5 from
their speaker-photo placeholders, changes their backgrounds from white to
pale yellow (\texttt{\#FFFFCC}), saves the presentation, and cross-checks the
live UNO state against the saved XML while confirming that the other four
slides remain unchanged.

\begin{figure}[htbp]
    \centering
    \includegraphics[width=\linewidth]{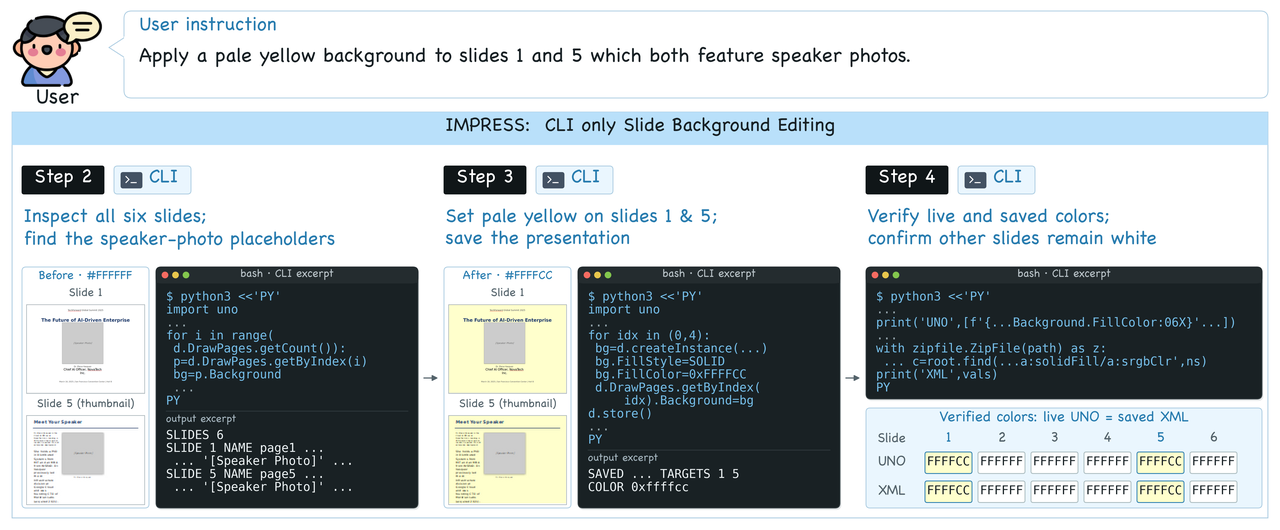}
    \caption{\textbf{A representative CLI only trajectory.}}
    \label{fig:cli_only_case}
\end{figure}

\paragraph{Interleaved GUI--CLI trajectories.}
Two construction routes produce trajectories containing both interfaces.
In the first, Qwen3.8-27B has access to GUI and CLI actions and chooses
between them during execution. In the second, we apply a semantics-preserving
action-level transformation to GUI only trajectories generated by Qwen3.8-27B. We first
identify typing actions that enter commands in a terminal, rather than
ordinary text in an application. Because one command may be split across
many actions, we reconstruct the complete command before replacement. For
example, typing \texttt{mkdir -p /tmp/proj} and then pressing Enter becomes
one direct CLI action; likewise, a heredoc entered line by line is folded
into one multiline command. We rewrite both the executable action and its
tool call, remove terminal-opening shortcuts, submission keystrokes, and
folded content lines, and preserve all genuine GUI interactions and
non-command text input. Thus, the transformation may fold multiple GUI
actions into one CLI action, but never splits one action into several. The
converted trajectories are replayed, and only successful replays are
retained, since a shorter command sequence need not preserve the application
state expected by subsequent GUI actions. Figure~\ref{fig:hybrid_case}
illustrates this interaction. The CLI first installs the Night Owl extension,
making it available to VS Code; GUI actions then open the theme picker,
select and preview the exact theme, and confirm the choice. Finally, the
trajectory returns to the CLI and reads back the persisted
\texttt{workbench.colorTheme} setting, verifying that the GUI selection was
saved. Thus, direct commands efficiently handle setup and state inspection,
while the GUI is retained for the visually grounded, stateful selection
step.

\begin{figure}[htbp]
    \centering
    \includegraphics[width=\linewidth]{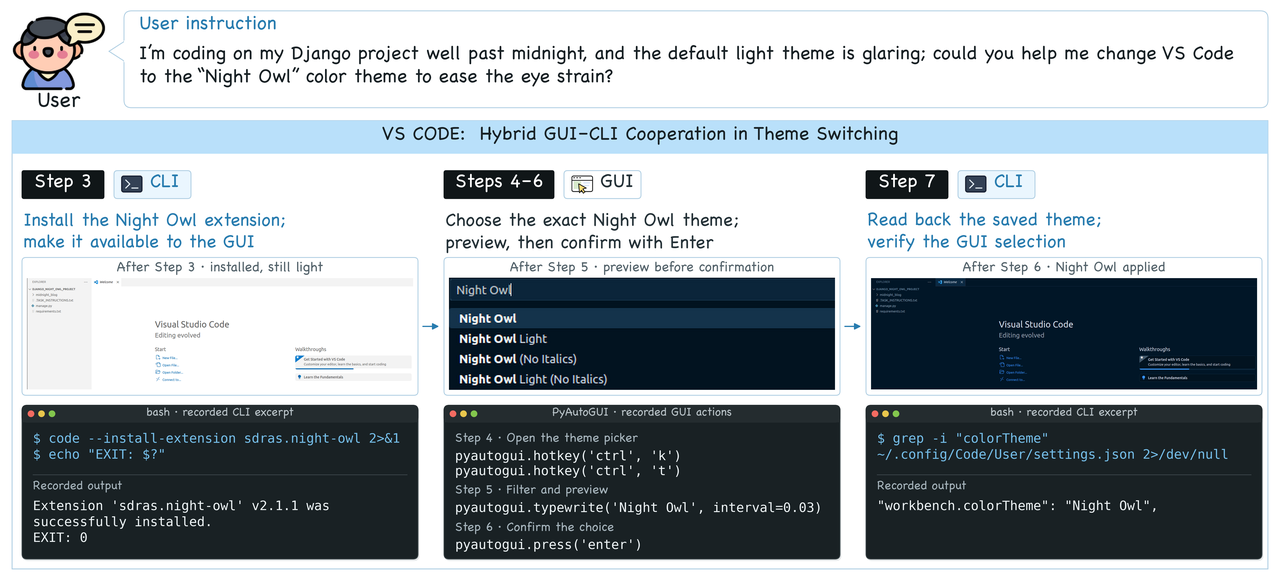}
    \caption{\textbf{A representative interleaved GUI--CLI trajectory.}
    CLI actions install the requested extension and verify the persisted
    setting, while GUI actions select, preview, and confirm the exact theme.}
    \label{fig:hybrid_case}
\end{figure}

\subsection{RLVR Task Construction and CLI Advantage Labeling}
\label{app:task_labels}

\paragraph{RLVR task construction.}
Building on CUA-Gym~\citep{wang2026cuagym}, we use application-specific
interface guides to synthesize task instructions, initial environment states,
required assets, and executable verifiers. We retain tasks with executable
verifiers and reachable target states, yielding 3{,}000 verified tasks across
11 domains (Table~\ref{tab:rlvr_domain_distribution}).

\begin{table}[!htbp]
    \centering
    \caption{Domain distribution of the full 3{,}000-task RLVR pool,
    before sampling the 1{,}000 tasks used for online RL.}
    \label{tab:rlvr_domain_distribution}
    \small
    \setlength{\tabcolsep}{12pt}
    \begin{tabular}{@{}lr@{}}
        \toprule
        \textbf{Domain} & \textbf{Share} \\
        \midrule
        \texttt{libreoffice\_calc} & 26.8\% \\
        \texttt{libreoffice\_writer} & 14.4\% \\
        \texttt{libreoffice\_impress} & 13.3\% \\
        \texttt{multi\_apps} & 12.9\% \\
        \texttt{os} & 8.9\% \\
        \texttt{chrome} & 5.4\% \\
        \texttt{vs\_code} & 4.4\% \\
        \texttt{pdf} & 4.3\% \\
        \texttt{vlc} & 3.6\% \\
        \texttt{gimp} & 3.1\% \\
        \texttt{thunderbird} & 2.9\% \\
        \midrule
        Total & 100.0\% \\
        \bottomrule
    \end{tabular}
\end{table}

\paragraph{Labeling procedure.}
For each of the 3{,}000 verified tasks, we sample 16 rollouts with
Qwen3.8-27B under each of GUI only, CLI only, and GUI--CLI action spaces.
Success rates use all 16 rollouts per mode, without filtering by CLI step
share. We rank modes by success rate, breaking ties within 5 percentage
points by the median step count of successful rollouts; residual ties favor
a single-interface mode, then GUI only. We set $b^\star=1$ if the top-ranked
mode is CLI only, or if it is GUI--CLI and more than half of its successful
rollouts contain at least one direct CLI command; otherwise, $b^\star=0$.
No task is removed during labeling.

\section{Training Details}
\label{app:training_details}

\subsection{Supervised Fine-Tuning Configuration}
\label{app:sft_configuration}

The training corpus consists of 5{,}023 successful computer-use trajectories
collected by driving a graphical desktop, each paired with a natural-language
task instruction. Expanding each trajectory at every decision step yields
52{,}227 step-level samples. Each sample is a demonstration prefix for a target
action $a_t$, with context comprising the task instruction and preceding
observations and actions. Filtering these samples to at most 12{,}000 tokens
yields 46{,}876 step-level samples.
Each prefix retains at most the three most recent screenshots at different
resolutions. The current observation, on which the target action operates,
is kept at higher resolution with a pixel budget of at most 2{,}088{,}960
pixels, costing 2{,}040 visual tokens. The two preceding screenshots are
downsampled by a factor of two along each dimension, with a configured pixel
budget of 548{,}800 pixels and a cost of 510 visual tokens each. Visual input
therefore contributes at most $2{,}040 + 2 \times 510 = 3{,}060$ tokens per
sample. The higher-resolution current frame preserves the spatial detail
needed for coordinate-level actions, while downsampled history supplies
semantic context. Older screenshots are replaced with the placeholder text
\texttt{This screenshot has been collapsed.}, and the action history is
limited to the 30 most recent steps. Across the corpus, 78.4\% of samples
contain three live screenshots, 10.8\% contain two, and 10.7\% contain one,
corresponding to an average of 2{,}895 visual tokens per sample and 45.1\% of
all processed tokens. With three screenshots, the 12{,}000-token context
budget leaves at most 8{,}940 tokens for the text prompt and target response;
the longest text prompt observed across the corpus contains 8{,}945 tokens.
Only the final assistant turn of each prefix contributes to the supervised
loss; all preceding assistant turns remain in the context but are masked.
The supervised target preserves the model's reasoning block.

The model is initialized from Qwen3.5-9B and trained with \texttt{verl} and
Megatron-LM for 2 epochs at a global batch size of 256 step-level samples,
yielding $\lfloor 46{,}876 / 256 \rfloor \times 2 = 366$ optimizer updates.
Sequences are padded to the right, batched dynamically with a
12{,}000-token budget per GPU, and truncated from the left if they exceed
the context limit. Training uses 2 nodes with 8 GPUs each; TP, PP, and DP
denote tensor, pipeline, and data parallelism, respectively. Validation is
not used for model selection, and the checkpoint at the final update is
reported. Table~\ref{tab:sft_hyperparameters} lists the full training
configuration.

\begin{table}[!t]
    \centering
    \caption{Supervised fine-tuning hyperparameters.}
    \label{tab:sft_hyperparameters}
    \small
    \setlength{\tabcolsep}{5pt}
    \renewcommand{\arraystretch}{1.0}
    \begin{tabular}{@{}>{\raggedright\arraybackslash}p{0.31\linewidth}
        >{\raggedright\arraybackslash}p{0.25\linewidth}
        >{\raggedright\arraybackslash}p{0.39\linewidth}@{}}
        \toprule
        \textbf{Category} & \textbf{Hyperparameter} & \textbf{Value} \\
        \midrule
        \rowcolor{gray!15}[0pt][0pt]
        \multicolumn{3}{@{}l@{}}{\textit{Model and Training Data}} \\
        Base model & Initialization & Qwen3.5-9B \\
        Training data & Samples / trajectories & 46{,}876 / 5{,}023 \\
        \midrule
        \rowcolor{gray!15}[0pt][0pt]
        \multicolumn{3}{@{}l@{}}{\textit{Context and Visual Input}} \\
        Context length & Tokens / truncation & 12{,}000 / left \\
        Screenshots & Current / history frames & 1 / 2 \\
        Visual tokens & Current / history frame & 2{,}040 / 510 \\
        Action history & Steps & 30 \\
        \midrule
        \rowcolor{gray!15}[0pt][0pt]
        \multicolumn{3}{@{}l@{}}{\textit{Batch Size and Training Duration}} \\
        Global batch size & Step-level samples & 256 \\
        Training duration & Epochs / updates & 2 / 366 \\
        \midrule
        \rowcolor{gray!15}[0pt][0pt]
        \multicolumn{3}{@{}l@{}}{\textit{Infrastructure and Parallelism}} \\
        Cluster & Nodes / GPUs per node & 2 / 8 \\
        Total GPUs & -- & 16 \\
        Parallelism & TP / PP / DP & 2 / 1 / 8 \\
        Hardware & GPU & NVIDIA H20 \\
        \midrule
        \rowcolor{gray!15}[0pt][0pt]
        \multicolumn{3}{@{}l@{}}{\textit{Optimization and Systems}} \\
        Learning rate & Initial / schedule / warm-up &
        $1\times10^{-5}$ / cosine / 10\% \\
        Optimizer & Type / $\beta_1$ / $\beta_2$ & AdamW / 0.9 / 0.999 \\
        Regularization & Weight decay / gradient clip & 0.01 / 1.0 \\
        Precision & -- & bfloat16 \\
        Attention backend & -- & FlashAttention \\
        Activation recomputation & Mode / layers & Full, uniform / 1 \\
        Optimizer state & Distribution / offload & Distributed / CPU \\
        Dynamic batching & Tokens per GPU & 12{,}000 \\
        \midrule
        \rowcolor{gray!15}[0pt][0pt]
        \multicolumn{3}{@{}l@{}}{\textit{Evaluation and Reproducibility}} \\
        Validation / checkpoint & -- & None / final update \\
        Random seed & -- & 1 \\
        \bottomrule
    \end{tabular}
\end{table}

\subsection{Online Reinforcement Learning Configuration}
\label{app:rl_configuration}

\paragraph{Tasks and training setup.}
We construct tasks with custom definitions, initial states, and success
criteria on an OSWorld-compatible desktop framework, rather than reuse public
GUI benchmark tasks. Starting from the SFT checkpoint, we train with GRPO on
1{,}000 tasks sampled from 3{,}000 verified tasks, using \texttt{slime} with
Megatron-LM for optimization and SGLang for rollout.
Table~\ref{tab:rl_hyperparameters} lists the configuration; its sampling
settings apply only to training rollouts, not benchmark evaluation.

\paragraph{Reward design and group filtering.}
The task evaluator provides a binary success reward $R_{\mathrm{acc}}$,
augmented by a success-gated modality-preference reward:
\begin{equation}
    R_{\mathrm{CLI}}(\tau)
    = \mathbb{I}\!\left[\mathrm{Success}(\tau)\right]
      \mathbb{I}\!\left[b(\tau)=b^\star\right].
    \label{eq:app_cli_preference}
\end{equation}
Here $b(\tau)\in\{0,1\}$ indicates whether $\tau$ issues at least one direct
CLI command and $b^\star\in\{0,1\}$ is the task's calibrated preference from
Appendix~\ref{app:task_labels}. Every task in the pool carries a label, so
this term is defined for all training tasks. We normalize the combined trajectory
reward $R(\tau)=R_{\mathrm{acc}}+\lambda_{\mathrm{CLI}}R_{\mathrm{CLI}}(\tau)$
within each GRPO group, with $\lambda_{\mathrm{CLI}}=0.1$. The execution term
$r_t^{\mathrm{exec}}$ in Eq.~(\ref{eq:exec_reward}) is added \emph{after}
normalization with $\lambda_{\mathrm{exec}}=0.3$, as in
Eq.~(\ref{eq:step_advantage}). Before normalization, we discard groups with
$\max_{\tau\in\mathcal{G}}R(\tau)-\min_{\tau\in\mathcal{G}}R(\tau)
\leq10^{-12}$, retaining at least one group to avoid empty batches. This
criterion uses the combined trajectory reward, not the binary outcome alone,
and excludes the later step-level execution term. Filtering removes groups
without a group-relative learning signal and makes the effective batch size
variable.

\paragraph{Context construction and step-level credit assignment.}
Dynamic history expands each trajectory into step-level samples, each
concatenating the context visible at that step with the policy-generated
response. At most three screenshots are retained, including the current
frame; historical frames are downsampled and older images are replaced by
text placeholders. Only the latest three steps are rendered in full, with
earlier steps reduced to one-line action summaries. Pixel budgets, output
truncation, and step limits are given in Table~\ref{tab:rl_hyperparameters}. Only the current response contributes to the loss; all context tokens are
masked, and samples without trainable tokens are skipped. Group statistics
and normalized advantages are computed after deduplicating step samples by
trajectory, then each trajectory's group-relative advantage is broadcast
back to its steps before adding the local execution term. This prevents
long trajectories from being counted repeatedly in group normalization.

\paragraph{Trajectory validity and failure handling.}
Aborted trajectories receive zero reward and are excluded from gradient
computation. Environment lifecycle failures are recorded with their stage
and error description in sample metadata. In-VM execution failures returned
by the executor rather than raised as exceptions do not by themselves
trigger sample removal; affected steps remain eligible for training, with
applicable CLI execution penalties reflected in $r_t^{\mathrm{exec}}$.

\paragraph{Asynchronous rollout and policy staleness.}
Optimization and rollout are decoupled to reduce idle time during environment
resets. Each step records its generating policy version, and a group's birth
version $v_{\mathrm{birth}}$ is the minimum over all its steps. We discard
groups when $v_{\mathrm{cur}}-v_{\mathrm{birth}}>\delta$, where
$v_{\mathrm{cur}}$ is the current policy version and $\delta=2$ optimizer
updates.

\begin{table}[!t]
    \centering
    \caption{Online reinforcement learning hyperparameters.}
    \label{tab:rl_hyperparameters}
    \small
    \setlength{\tabcolsep}{5pt}
    \renewcommand{\arraystretch}{1.0}
    \begin{tabular}{@{}>{\raggedright\arraybackslash}p{0.31\linewidth}
        >{\raggedright\arraybackslash}p{0.25\linewidth}
        >{\raggedright\arraybackslash}p{0.39\linewidth}@{}}
        \toprule
        \textbf{Category} & \textbf{Hyperparameter} & \textbf{Value} \\
        \midrule
        \rowcolor{gray!15}[0pt][0pt]
        \multicolumn{3}{@{}l@{}}{\textit{Tasks and Infrastructure}} \\
        Training tasks & Sampled / verified & 1{,}000 / 3{,}000 \\
        Training stack & Framework / optimizer / rollout &
        \texttt{slime} / Megatron-LM / SGLang \\
        Cluster & Nodes / GPUs per node & 3 / 8 \\
        Total GPUs & -- & 24 \\
        Hardware & GPU & NVIDIA H20 \\
        \midrule
        \rowcolor{gray!15}[0pt][0pt]
        \multicolumn{3}{@{}l@{}}{\textit{Parallelism and Batch Size}} \\
        Training & GPUs / TP / DP & 16 / 4 / 4 \\
        Inference & GPUs / engines & 8 / 8 \\
        Rollouts per prompt & -- & 8 \\
        Nominal batch & Groups / trajectories & 8 / 64 \\
        \midrule
        \rowcolor{gray!15}[0pt][0pt]
        \multicolumn{3}{@{}l@{}}{\textit{Rewards and GRPO}} \\
        Outcome reward & $R_{\mathrm{acc}}$ & $\{0,1\}$ \\
        CLI preference & $\lambda_{\mathrm{CLI}}$ / level & 0.1 / trajectory \\
        Execution term & $\lambda_{\mathrm{exec}}$ / level & 0.3 / step \\
        Clip ratio & Lower / upper & 0.2 / 0.2 \\
        KL & Penalty / loss / estimator & 0.001 / 0.01 / k3 \\
        Reference / entropy & -- & SFT checkpoint / 0 \\
        \midrule
        \rowcolor{gray!15}[0pt][0pt]
        \multicolumn{3}{@{}l@{}}{\textit{Context and Sequence Length}} \\
        Screenshots & Maximum frames & 3 \\
        Full-text history & Steps & 3 \\
        Image pixels & Current / historical & 2{,}088{,}960 / 548{,}800 \\
        CLI output limit & Characters per step & 1{,}000 \\
        Response limit & Tokens per turn & 4{,}096 \\
        Environment limit & Training / evaluation steps & 30 / 50 \\
        \midrule
        \rowcolor{gray!15}[0pt][0pt]
        \multicolumn{3}{@{}l@{}}{\textit{Optimization}} \\
        Learning rate & Schedule / warm-up & $1\times10^{-6}$ / constant / 0 \\
        Optimizer & $\beta_1$ / $\beta_2$ / weight decay & Adam / 0.9 / 0.95 / 0.1 \\
        Gradient & Clipping / microbatch & 1.0 / 1 \\
        Checkpoint interval & Updates & 10 \\
        \midrule
        \rowcolor{gray!15}[0pt][0pt]
        \multicolumn{3}{@{}l@{}}{\textit{Rollout Engine}} \\
        Temperature / top-$p$ & -- & 1.0 / 1.0 \\
        Thinking / KV cache & -- & Disabled / 0.7 \\
        Concurrency & Requests per engine & 64 \\
        Chunked prefill & Tokens & 4{,}096 \\
        Reset / post-action wait & Seconds & 60 / 0.5 \\
        Maximum policy lag & $\delta$ (updates) & 2 \\
        \bottomrule
    \end{tabular}
\end{table}

\section{Evaluation Protocols and Metrics}
\label{app:evaluation_protocol}
\label{app:benchmark_settings}
\label{app:evaluation_metrics}

\paragraph{Evaluated task sets.}
OSWorld~\citep{xie2024osworld} is the primary benchmark and we evaluate its
full set of $N=361$ tasks. For out-of-distribution transfer we evaluate the
$361$ tasks of OSWorld-MCP~\citep{jia2025osworldmcp}, which shares the
OSWorld task set but exposes an MCP-enabled environment, and the $154$ tasks
of WindowsAgentArena~\citep{bonatti2024windowsagentarena}, which changes the
operating system. Because the two transfer settings vary different aspects of
the environment, and because OSWorld-MCP is built on the same task set as
OSWorld rather than on unseen tasks, they are reported separately and are not
combined into a single aggregate score. Every model is given a budget of at
most $50$ environment steps per task, and each reported number comes from a
\emph{single} evaluation run per model and benchmark.

\paragraph{Reported metrics.}
Let $v_i\in[0,1]$ denote the evaluator score of task $i$. The reported
\textbf{Acc.} is the mean evaluator score over the evaluated task set,
\begin{equation}
    \mathrm{Acc.}=\frac{100}{N}\sum_{i=1}^{N}v_i .
    \label{eq:app_accuracy}
\end{equation}
Most OSWorld verifiers are binary, in which case $v_i\in\{0,1\}$ and
Eq.~(\ref{eq:app_accuracy}) reduces to a strict success rate; a subset of
verifiers, however, returns a fractional score, and we average these scores
as returned rather than thresholding them into binary outcomes. The reported
values are therefore mean evaluator scores, which are an upper bound on the
strict all-or-nothing success rate and should be compared only against
numbers aggregated the same way. A model-generated \texttt{terminate}
declaration never contributes to $v_i$.
Let $T_i$ further denote the number of environment interaction steps consumed
by task $i$. The reported \textbf{Avg. Steps} is the all-task mean
$\overline{T}_{\mathrm{all}}=N^{-1}\sum_{i=1}^{N}T_i$, computed over the same
task set as Eq.~(\ref{eq:app_accuracy}) rather than over successful tasks
only. One step is one environment interaction, so \texttt{wait} and the
terminating action are counted, and a task that exhausts the budget
contributes $T_i=50$.

\paragraph{Interface access.}
Interface access is part of the evaluation configuration, not solely a
property of the model name: a GUI only agent, a GUI--API agent, and a
GUI--CLI agent may have different executable capabilities on the same task.
In particular, evaluating in an MCP-enabled environment does not by itself
establish that every model receives identical MCP tools, and
cross-operating-system evaluation requires an explicit description of the
Windows command-execution interface rather than an assumption that a Linux
shell configuration transfers unchanged.

Let $N_{\mathrm{CLI}}$ and $N_{\mathrm{GUI}}$ count executable steps assigned
to the two interfaces in a specified rollout set. The pooled CLI step share
is $100N_{\mathrm{CLI}}/(N_{\mathrm{CLI}}+N_{\mathrm{GUI}})$; control actions
are not executable GUI or CLI steps. The CLI execution error rate is
$100N_{\mathrm{CLI,err}}/N_{\mathrm{CLI}}$, using the same failure predicate
as $r_t^{\mathrm{exec}}$. A rate with a zero denominator is undefined, not zero. Pooled step
shares weight longer trajectories more heavily and are distinct from an
unweighted mean of per-task shares.

\section{More Case Studies}
\label{app:qualitative_analysis}
\label{app:cooperation_cases}

\begingroup
\raggedbottom
\newcounter{hybridcase}
\definecolor{CaseInk}{HTML}{243447}
\definecolor{CaseMuted}{HTML}{63758A}
\definecolor{CaseGUI}{HTML}{4676AD}
\definecolor{CaseCLI}{HTML}{258078}
\definecolor{CaseControl}{HTML}{7B7295}
\definecolor{CasePaper}{HTML}{F5F7FA}
\definecolor{CaseLine}{HTML}{D6DFE8}

\lstdefinestyle{caseaction}{
    basicstyle=\fontencoding{T1}\fontfamily{lmtt}\fontsize{7.2}{8.2}\selectfont,
    keywordstyle=\color{CaseInk},
    commentstyle=\color{CaseMuted},
    stringstyle=\color{CaseInk},
    columns=fullflexible,
    keepspaces=true,
    showstringspaces=false,
    breaklines=true,
    breakatwhitespace=false,
    breakindent=8pt,
    aboveskip=3pt,
    belowskip=0pt,
    tabsize=4,
    upquote=true,
    literate={-}{{\char45}}1,
    numbers=none
}
\lstset{style=caseaction}
\lstdefinestyle{caseoutput}{
    style=caseaction,
    basicstyle=\fontencoding{T1}\fontfamily{lmtt}\fontsize{7.2}{8.2}\selectfont\color{CaseInk},
    aboveskip=2pt
}
\newcommand{\caseoutput}[1]{%
    \par\vspace{4pt}%
    {\color{CaseCLI!35!white}\hrule height0.4pt}%
    \vspace{3pt}%
    {\sffamily\fontsize{6.5}{8}\selectfont\color{CaseCLI}%
        EXECUTION RESULT\hfill\texttt{#1}\par}%
}

\newenvironment{caseoverview}[4]{%
    \begin{tcolorbox}[
        enhanced,
        colback=CasePaper,
        colframe=CaseInk,
        colbacktitle=CaseInk,
        coltitle=white,
        boxrule=0.5pt,
        arc=2pt,
        left=8pt,right=8pt,top=7pt,bottom=7pt,
        toptitle=5pt,bottomtitle=5pt,
        before skip=8pt,after skip=8pt,
        fonttitle=\sffamily\bfseries\small,
        title={Case #1\quad #2}]
    \fontsize{9}{11}\selectfont\raggedright
    \textcolor{CaseCLI}{\sffamily\bfseries Recorded score: 1.0}%
    \hfill{\sffamily #3 steps}\par
    {\fontsize{7.5}{9}\selectfont\textcolor{CaseMuted}{Task ID: \texttt{#4}}}\par\smallskip
}{%
    \end{tcolorbox}%
}

\newenvironment{casestep}[4]{%
    \begin{tcolorbox}[
        enhanced,
        sidebyside,
        sidebyside align=top,
        lefthand ratio=0.43,
        sidebyside gap=10pt,
        colback=Case#2!5!white,
        colframe=Case#2!45!white,
        colbacktitle=Case#2!12!white,
        coltitle=Case#2!75!black,
        boxrule=0.5pt,
        arc=2pt,
        segmentation style={solid,draw=Case#2!20!white,line width=0.4pt},
        left=6pt,right=6pt,top=3pt,bottom=3pt,
        toptitle=3pt,bottomtitle=3pt,
        before skip=3pt,after skip=3pt,
        fonttitle=\sffamily\fontsize{8}{10}\selectfont,
        title={\textbf{STEP #1 / \casetotal}\quad\textbf{#2}%
            \hfill Case \casenumber\enspace\textbar\enspace\casename}]
    {\sffamily\fontsize{6.5}{8}\selectfont\color{CaseMuted}GUI OBSERVATION\par}
    \vspace{3pt}
    \includegraphics[width=\linewidth]{\casepath/#3}\par
    \tcblower
    {\sffamily\fontsize{6.5}{8}\selectfont\color{CaseMuted}MODEL ACTION\quad\texttt{#4}\par}
}{%
    \end{tcolorbox}%
}

We present three complete recorded trajectories, with 8, 9, and 8 steps,
respectively. All three have a saved evaluator score of 1.0. Each row pairs
\textbf{the pre-action screenshot on the left} with \textbf{the model action
on the right}; the next row shows the resulting observation. Blue, teal,
and lavender distinguish GUI, CLI, and control actions. The right column
reproduces each \texttt{bash} command in full, or the parameters of a control
action, with only typographic line wrapping. Each CLI action is followed
by its recorded execution status and stdout; all recorded stderr fields
contain only whitespace and are omitted. Trailing blank lines are trimmed,
but diagnostic messages are retained. Reasoning is omitted. Waiting and
termination steps are retained; execution status and the agent's termination
declaration are distinct from the saved task-evaluator score.

\def\casenumber{1}
\def\casetotal{8}
\def\casename{Vim line numbers}
\def\casepath{figures/case_studies/vim_linenumber}

\refstepcounter{hybridcase}
\begin{caseoverview}{1}{Web search to persistent Vim configuration}{8}{b337d106-053f-4d37-8da0-7f9c4043a66b}
\textbf{Task.} Search the Internet for a tutorial on absolute line numbers
in Vim and make them the default for the local editor.\par\smallskip
\textbf{Hybrid pattern: Context handoff.}
GUI information gathering $\rightarrow$ CLI configuration.\par\smallskip
\textcolor{CaseGUI}{\textbf{GUI for information acquisition.}}
Steps 1--4 search the web and inspect the displayed instructions, which
identify \texttt{set number} and the \texttt{.vimrc} configuration file.\par\smallskip
\textcolor{CaseCLI}{\textbf{CLI for persistence and verification.}}
Steps 6--7 append the setting to \texttt{\string~/.vimrc} and read back its
contents.\par\smallskip
\textbf{Analysis.} The handoff connects an instruction observed in the
browser to a concrete local configuration change. The two interfaces
address different parts of the task: GUI browsing supplies the requested
online guidance, while direct file editing makes the setting persistent
rather than enabling it only for one Vim session. The recorded read-back
confirms \texttt{set number}, and the evaluator score is 1.0; no subsequent
Vim launch is shown.
\end{caseoverview}
\label{app:case_vim}

\begin{casestep}{01}{GUI}{step_0.png}{bash}
\begin{lstlisting}
python3 <<'PY'
import pyautogui
# Click on the Chrome browser window to bring it to focus
pyautogui.click(700, 275)
PY
\end{lstlisting}
\end{casestep}

\begin{casestep}{02}{GUI}{step_1.png}{bash}
\begin{lstlisting}
python3 <<'PY'
import pyautogui
# Click on the Google search bar
pyautogui.click(517, 367)
# Type the search query
pyautogui.typewrite('tutorial adding absolute line numbers Vim set number as default', interval=0.02)
PY
\end{lstlisting}
\end{casestep}

\begin{casestep}{03}{GUI}{step_2.png}{bash}
\begin{lstlisting}
python3 <<'PY'
import pyautogui
pyautogui.press('enter')
PY
\end{lstlisting}
\end{casestep}

\begin{casestep}{04}{GUI}{step_3.png}{bash}
\begin{lstlisting}
python3 <<'PY'
import pyautogui
# Scroll down to see more of the tutorial
pyautogui.scroll(-5)
PY
\end{lstlisting}
\end{casestep}

\begin{casestep}{05}{GUI}{step_4.png}{bash}
\begin{lstlisting}
python3 <<'PY'
import pyautogui
# Click on the terminal window to bring it to focus
pyautogui.click(250, 275)
PY
\end{lstlisting}
\end{casestep}

\begin{casestep}{06}{CLI}{step_5.png}{bash}
\begin{lstlisting}
echo "set number" >> ~/.vimrc && echo "Added to .vimrc. Contents:" && cat ~/.vimrc
\end{lstlisting}
\caseoutput{success}
\begin{lstlisting}[style=caseoutput]
Added to .vimrc. Contents:
set number
\end{lstlisting}
\end{casestep}

\begin{casestep}{07}{CLI}{step_6.png}{bash}
\begin{lstlisting}
echo "=== .vimrc location and content ===" && ls -la ~/.vimrc && echo "" && echo "=== Content ===" && cat -A ~/.vimrc | head
\end{lstlisting}
\caseoutput{success}
\begin{lstlisting}[style=caseoutput]
=== .vimrc location and content ===
-rw-r--r-- 1 user user 11 Sep  7 15:45 /home/user/.vimrc

=== Content ===
set number$
\end{lstlisting}
\end{casestep}

\begin{casestep}{08}{Control}{step_7.png}{terminate}
\begin{lstlisting}
terminate(status="success")
\end{lstlisting}
\end{casestep}

\par\medskip
\def\casenumber{2}
\def\casetotal{9}
\def\casename{VLC volume limit}
\def\casepath{figures/case_studies/vlc_maxvolume}

\refstepcounter{hybridcase}
\begin{caseoverview}{2}{GUI preferences to a direct VLC configuration edit}{9}{9195653c-f4aa-453d-aa95-787f6ccfaae9}
\textbf{Task.} Raise VLC's maximum volume from 125\% to 200\%.\par\smallskip
\textbf{Hybrid pattern: Adaptive interface fallback.}
GUI exploration $\rightarrow$ CLI configuration repair.\par\smallskip
\textcolor{CaseGUI}{\textbf{GUI for preference exploration.}}
Steps 1--6 navigate VLC's preferences and audio settings, including a wait;
this sequence does not establish the requested maximum.\par\smallskip
\textcolor{CaseCLI}{\textbf{CLI for targeted configuration and read-back.}}
Step 7 locates \texttt{vlcrc} and finds \texttt{\#qt-max-volume=125}.
Step 8 replaces it with \texttt{qt-max-volume=200} and checks the saved
value.\par\smallskip
\textbf{Analysis.} Instead of continuing the same GUI search, the agent
changes the level of interaction from a preferences dialog to its backing
configuration file. This is a fallback from an unproductive interaction
route, not evidence that VLC's GUI cannot support the setting. The CLI
completes the change and exposes an explicit value for verification.
The evaluator relaunches VLC and checks the configuration, returning 1.0;
the rollout itself does not show the updated volume control.
\end{caseoverview}
\label{app:case_vlc}

\begin{casestep}{01}{GUI}{step_0.png}{bash}
\begin{lstlisting}
python3 <<'PY'
import pyautogui
pyautogui.click(493, 344)
PY
\end{lstlisting}
\end{casestep}

\begin{casestep}{02}{GUI}{step_1.png}{bash}
\begin{lstlisting}
python3 <<'PY'
import pyautogui
pyautogui.click(566, 557)
PY
\end{lstlisting}
\end{casestep}

\begin{casestep}{03}{GUI}{step_2.png}{bash}
\begin{lstlisting}
python3 <<'PY'
import pyautogui
pyautogui.click(355, 248)
PY
\end{lstlisting}
\end{casestep}

\begin{casestep}{04}{GUI}{step_3.png}{bash}
\begin{lstlisting}
python3 <<'PY'
import pyautogui
pyautogui.click(331, 831)
PY
\end{lstlisting}
\end{casestep}

\begin{casestep}{05}{Control}{step_4.png}{wait}
\begin{lstlisting}
wait(time=1)
\end{lstlisting}
\end{casestep}

\begin{casestep}{06}{GUI}{step_5.png}{bash}
\begin{lstlisting}
python3 <<'PY'
import pyautogui
pyautogui.click(355, 248)
PY
\end{lstlisting}
\end{casestep}

\begin{casestep}{07}{CLI}{step_6.png}{bash}
\begin{lstlisting}
find / -name "vlcrc" 2>/dev/null; echo "---"; grep -n "max-volume" ~/.config/vlc/vlcrc 2>/dev/null || echo "not in home"
\end{lstlisting}
\caseoutput{success}
\begin{lstlisting}[style=caseoutput]
/home/user/.config/vlc/vlcrc
---
2650:#qt-max-volume=125
\end{lstlisting}
\end{casestep}

\begin{casestep}{08}{CLI}{step_7.png}{bash}
\begin{lstlisting}
sed -i 's/^#qt-max-volume=125/qt-max-volume=200/' /home/user/.config/vlc/vlcrc && grep -n "qt-max-volume" /home/user/.config/vlc/vlcrc
\end{lstlisting}
\caseoutput{success}
\begin{lstlisting}[style=caseoutput]
2650:qt-max-volume=200
\end{lstlisting}
\end{casestep}

\begin{casestep}{09}{Control}{step_8.png}{terminate}
\begin{lstlisting}
terminate(status="success")
\end{lstlisting}
\end{casestep}

\par\medskip
\def\casenumber{3}
\def\casetotal{8}
\def\casename{Terminal screenshot}
\def\casepath{figures/case_studies/terminal_screenshot}

\refstepcounter{hybridcase}
\begin{caseoverview}{3}{CLI setup, GUI interaction, and screenshot verification}{8}{02ce9a50-7af2-47ed-8596-af0c230501f8}
\textbf{Task.} Run \texttt{ls} in \texttt{/home/user} for a Linux tutorial
and save a screenshot of the terminal as \texttt{ls.png} on the Desktop.\par\smallskip
\textbf{Hybrid pattern: Execution--verification loop.}
CLI setup $\rightarrow$ GUI execution $\rightarrow$ CLI capture and checks.\par\smallskip
\textcolor{CaseGUI}{\textbf{GUI for visible terminal execution.}}
Step 3 focuses the opened terminal, types \texttt{ls /home/user}, and
presses Enter so the output appears in the window to be captured.\par\smallskip
\textcolor{CaseCLI}{\textbf{CLI for setup, capture, and artifact inspection.}}
Steps 1--2 inspect the environment and open the terminal; steps 4--7 save
its screenshot, inspect the PNG, and open it in an image viewer.\par\smallskip
\textbf{Analysis.} A command result in the execution channel alone would
not provide the requested terminal image. Here, CLI setup creates the
window, GUI input produces the visible content, and CLI capture turns that
state into a saved artifact. The final observation displays the image
opened by step 7, closing the loop from file checks back to visual
inspection. Capture itself is a CLI action, not a GUI action. The recorded
PNG is $876\times533$ and the score is 1.0; the evaluator checks for
\texttt{ls} (or an OCR variant), not every directory entry.
\end{caseoverview}
\label{app:case_terminal_screenshot}

\begin{casestep}{01}{CLI}{step_0.png}{bash}
\begin{lstlisting}
which gnome-terminal xdg-open; ls ~/Desktop/ 2>/dev/null; echo "---"; ls /home/user | head -30
\end{lstlisting}
\caseoutput{success}
\begin{lstlisting}[style=caseoutput]
/usr/bin/gnome-terminal
/usr/bin/xdg-open
top-10-linux-commands-for-newbies.docx
---
Desktop
Documents
Downloads
Music
Pictures
Public
requirement.txt
server
snap
Templates
Videos
\end{lstlisting}
\end{casestep}

\begin{casestep}{02}{CLI}{step_1.png}{bash}
\begin{lstlisting}
(DISPLAY=:0 gnome-terminal --geometry=90x24 & sleep 3; echo "opened")
\end{lstlisting}
\caseoutput{success}
\begin{lstlisting}[style=caseoutput]
opened
\end{lstlisting}
\end{casestep}

\begin{casestep}{03}{GUI}{step_2.png}{bash}
\begin{lstlisting}
python3 <<'PY'
import pyautogui
pyautogui.click(265, 275)
import time
time.sleep(0.5)
pyautogui.typewrite('ls /home/user', interval=0.03)
time.sleep(0.3)
pyautogui.press('enter')
PY
\end{lstlisting}
\end{casestep}

\begin{casestep}{04}{CLI}{step_3.png}{bash}
\begin{lstlisting}
echo "=== windows ==="; wmctrl -lG 2>/dev/null | grep -iE "terminal|gnome" || echo "no wmctrl"; echo "=== tools ==="; which import gnome-screenshot scrot 2>/dev/null; echo "=== try gnome-screenshot -w ==="; DISPLAY=:0 gnome-screenshot -w -f ~/Desktop/ls.png 2>&1; echo "exit=$?"; ls -la ~/Desktop/ls.png 2>&1
\end{lstlisting}
\caseoutput{success}
\begin{lstlisting}[style=caseoutput]
=== windows ===
no wmctrl
=== tools ===
/usr/bin/gnome-screenshot
=== try gnome-screenshot -w ===
exit=0
-rw-r--r-- 1 user user 27955 Sep  7 15:47 /home/user/Desktop/ls.png
\end{lstlisting}
\end{casestep}

\begin{casestep}{05}{CLI}{step_4.png}{bash}
\begin{lstlisting}
file ~/Desktop/ls.png; echo "=== dimensions ==="; python3 -c "from PIL import Image; im=Image.open('/home/user/Desktop/ls.png'); print('size:', im.size, 'mode:', im.mode)" 2>/dev/null || echo "PIL not available"; echo "=== check for terminal text in image ==="; python3 -c "
from PIL import Image
im=Image.open('/home/user/Desktop/ls.png')
# sample a region where terminal text should be
w,h=im.size
print('image size', w, h)
"
\end{lstlisting}
\caseoutput{success}
\begin{lstlisting}[style=caseoutput]
/home/user/Desktop/ls.png: PNG image data, 876 x 533, 8-bit/color RGBA, non-interlaced
=== dimensions ===
size: (876, 533) mode: RGBA
=== check for terminal text in image ===
image size 876 533
\end{lstlisting}
\end{casestep}

\begin{casestep}{06}{CLI}{step_5.png}{bash}
\begin{lstlisting}
echo "=== terminal window geometry ==="; DISPLAY=:0 xdotool search --name "user@user-virtual-machine" getwindowgeometry %
from PIL import Image
im = Image.open('/home/user/Desktop/ls.png').convert('RGB')
w,h = im.size
px = im.load()
# count dark pixels (background) vs light pixels (text)
dark, light, mid = 0,0,0
for y in range(0,h,2):
    for x in range(0,w,2):
        r,g,b = px[x,y]
        v = (r+g+b)/3
        if v < 60: dark+=1
        elif v > 200: light+=1
        else: mid+=1
print(f"image {w}x{h}")
print(f"dark px: {dark}, light px: {light}, mid: {mid}")
# sample a horizontal strip in the middle to see text lines
for y in [h//2, h//2+20, h//2+40]:
    vals=[]
    for x in range(0,w,4):
        v=(px[x,y][0]+px[x,y][1]+px[x,y][2])/3
        vals.append(v)
    print(f"y={y}: min={min(vals):.0f} max={max(vals):.0f} mean={sum(vals)/len(vals):.0f}")
PY
\end{lstlisting}
\caseoutput{success}
\begin{lstlisting}[style=caseoutput]
=== terminal window geometry ===
=== analyze ls.png: dark bg + text ===
image 876x533
dark px: 114584, light px: 324, mid: 2038
y=266: min=0 max=166 mean=30
y=286: min=0 max=166 mean=30
y=306: min=0 max=166 mean=30
\end{lstlisting}
\end{casestep}

\begin{casestep}{07}{CLI}{step_6.png}{bash}
\begin{lstlisting}
(DISPLAY=:0 eog ~/Desktop/ls.png >/dev/null 2>&1 & sleep 2; echo "opened eog")
\end{lstlisting}
\caseoutput{success}
\begin{lstlisting}[style=caseoutput]
opened eog
\end{lstlisting}
\end{casestep}

\begin{casestep}{08}{Control}{step_7.png}{terminate}
\begin{lstlisting}
terminate(status="success")
\end{lstlisting}
\end{casestep}
\par\medskip
\endgroup

\begingroup
\renewcommand{\encodingdefault}{T1}
\fontencoding{T1}\selectfont
\newsavebox{\evaluationcardbox}
\newsavebox{\evaluationcodebox}
\renewcommand{\ttfamily}{\fontencoding{OT1}\fontfamily{cmtt}\selectfont}
\newsavebox{\promptpagebox}
\newsavebox{\prompttitlebox}
\newsavebox{\promptemptybox}
\newdimen\promptframewidth
\newdimen\promptremaining
\newcommand{\prompttypeset}{%
    \promptremaining=\dimexpr\pagegoal-\pagetotal-1pt\relax
    \ifdim\pagegoal=\maxdimen
        \promptremaining=\dimexpr\textheight-1pt\relax
    \fi
    \ifdim\promptremaining<60pt
        \newpage
        \promptremaining=\dimexpr\textheight-1pt\relax
    \fi
    \advance\promptremaining by -\ht\prompttitlebox
    \advance\promptremaining by -\dp\prompttitlebox
    \advance\promptremaining by -1pt
    \ifdim\dimexpr\ht\evaluationcardbox+\dp\evaluationcardbox>\promptremaining
        \setbox\promptpagebox=\vsplit\evaluationcardbox to \promptremaining
        \setbox\promptpagebox=\vbox{\unvbox\promptpagebox}%
        \ifdim\wd\evaluationcardbox=0pt
            \setbox\evaluationcardbox=\box\promptemptybox
            \let\promptnext\relax
        \else
            \def\promptnext{\newpage\prompttypeset}%
        \fi
    \else
        \setbox\promptpagebox=\box\evaluationcardbox
        \let\promptnext\relax
    \fi
    \nointerlineskip
    \hbox{\vbox{%
        \offinterlineskip
        \ifvoid\prompttitlebox\else
            {\color{black!65}\hrule height0.5pt}%
            \hbox{\color{black!65}\vrule width0.5pt
                \box\prompttitlebox\vrule width0.5pt}%
        \fi
        \hbox{\color{black!65}\vrule width0.5pt
            \kern8pt{\color{black}\box\promptpagebox}\kern8pt\vrule width0.5pt}%
        \ifvoid\evaluationcardbox
            {\color{black!65}\hrule height0.5pt}%
        \fi
    }}%
    \promptnext
}
\newenvironment{templatebox}[1]{%
    \par\addvspace{\medskipamount}%
    \begingroup
    \promptframewidth=\linewidth
    \setlength{\fboxsep}{0pt}%
    \setbox\prompttitlebox=\hbox{%
        \colorbox{black!80}{%
            \parbox{\dimexpr\promptframewidth-1pt\relax}{%
                \vspace{4pt}%
                \hspace*{8pt}{\color{white}\fontsize{11}{13}\selectfont
                    \rmfamily\bfseries #1}\par
                \vspace{4pt}}}}%
    \splittopskip=10pt
    \splitmaxdepth=0pt
    \vbadness=10000
    \setbox\evaluationcardbox=\vbox\bgroup
    \hsize=\dimexpr\promptframewidth-17pt\relax
    \linewidth=\hsize
    \columnwidth=\hsize
    \fontsize{10}{12}\selectfont\rmfamily
    \setlength{\parindent}{0pt}%
    \setlength{\parskip}{3pt}%
    \raggedright\frenchspacing
    \vskip5pt
}{%
    \par\nobreak\vskip6pt
    \egroup
    \prompttypeset
    \endgroup
    \par\addvspace{\medskipamount}%
}
\newcommand{\promptsection}[1]{%
    \par\addvspace{5pt}%
    {\fontsize{10}{12}\selectfont\sffamily\bfseries\#~#1\par}%
    \nobreak
}
\newsavebox{\promptcodepagebox}
\newcommand{\promptcoderow}[1]{%
    \nointerlineskip
    \hbox{\colorbox{black!2}{\hbox to \linewidth{%
        \color{black!22}\vrule width0.4pt\kern5pt
        {\color{black}#1}\hfil\kern5pt\vrule width0.4pt}}}%
}
\newcommand{\promptcodelines}{%
    \ifdim\dimexpr\ht\evaluationcodebox+\dp\evaluationcodebox>0pt
        \setbox\promptcodepagebox=\vsplit\evaluationcodebox to 9.2pt
        \setbox\promptcodepagebox=\vbox{\unvbox\promptcodepagebox}%
        \promptcoderow{\vrule width0pt height6.8pt depth2.4pt
            \box\promptcodepagebox}%
        \penalty0
        \expandafter\promptcodelines
    \fi
}
\newenvironment{evaluationcode}{%
    \par\begingroup
    \setlength{\fboxsep}{0pt}%
    \splittopskip=0pt
    \splitmaxdepth=2.4pt
    \setbox\evaluationcodebox=\vbox\bgroup
    \hsize=\dimexpr\linewidth-10.8pt\relax
    \linewidth=\hsize
    \fontsize{8}{9.2}\selectfont\ttfamily
    \setlength{\parskip}{0pt}%
    \setlength{\topsep}{0pt}%
    \setlength{\partopsep}{0pt}%
}{%
    \par\egroup
    \nobreak\vskip3pt
    {\color{black!22}\hrule height0.4pt}%
    \promptcoderow{\vrule width0pt height5pt}%
    \nobreak
    \promptcodelines
    \nobreak
    \promptcoderow{\vrule width0pt height5pt}%
    {\color{black!22}\hrule height0.4pt}%
    \vskip3pt
    \endgroup
}
\newenvironment{evaluationitems}{%
    \begin{list}{\textbullet}{%
        \setlength{\leftmargin}{12pt}%
        \setlength{\labelwidth}{6pt}%
        \setlength{\labelsep}{6pt}%
        \setlength{\itemsep}{3pt}%
        \setlength{\parsep}{0pt}%
        \setlength{\topsep}{2pt}%
        \setlength{\partopsep}{0pt}}
}{\end{list}}

\clearpage
\section{Prompt Used in Evaluation And Training Data Construction}
\label{app:evaluation_prompt}

\subsection{OSWorld Evaluation: Unified Action Schema}
\label{app:osworld_system_prompt}

This is the system prompt used for our OSWorld evaluation. A single \texttt{computer\_use} tool exposes
\texttt{action=bash} for both direct shell commands and GUI operations via
PyAutoGUI heredocs. It differs from the separate-tool \texttt{hybrid}
configuration in the action-schema analysis
(Table~\ref{tab:action_schema_analysis}), whose prompt is given in
Appendix~\ref{app:hybrid_system_prompt}.

\begin{templatebox}{System Prompt: OSWorld Evaluation (Unified Bash)}
\promptsection{Role}
You are a multi-purpose intelligent assistant operating a computer through a bash terminal.
The password of the computer is \texttt{password}.

\promptsection{Environment}
You face a machine with a graphical desktop AND a bash terminal. You act ONLY
by running one shell command per step (\texttt{action=bash}): write shell for
CLI/file operations, or drive the GUI with \texttt{pyautogui} inside a quoted
heredoc \texttt{python3 <<'PY' ... PY} (coordinates are 0--999). Each step you
are shown the latest screenshot plus the previous command's output.

\promptsection{Tools}
You have access to the following functions:
\begin{evaluationcode}
\begin{verbatim}
Tool type:             function
Function name:         computer_use
Parameters type:       object
Required parameters:   ["action"]
\end{verbatim}
\end{evaluationcode}
Control the machine through a bash terminal. One action per call. The screen
is a GUI desktop: drive it with \texttt{pyautogui} via a quoted heredoc, or run
shell commands / file operations directly.

\promptsection{Tool Parameters}
\begin{evaluationitems}
    \item \textbf{\texttt{action}} \textit{(string; required).}
    One of \texttt{bash}, \texttt{wait}, \texttt{terminate}, \texttt{answer}.
    The action descriptions are given below.
    \item \textbf{\texttt{command}} \textit{(string).}
    Required by \texttt{action=bash}. A single shell command. For GUI, run
    \texttt{pyautogui} via a quoted heredoc:
    \texttt{python3 <<'PY' ... PY} (coordinates 0--999).
    \item \textbf{\texttt{timeout}} \textit{(number).}
    Optional for \texttt{action=bash}: seconds to allow the command
    (default 60). The sandbox cuts commands short after approximately
    30\,s regardless.
    \item \textbf{\texttt{time}} \textit{(number).}
    Optional for \texttt{action=wait}: seconds.
    \item \textbf{\texttt{status}} \textit{(string).}
    One of \texttt{success}, \texttt{failure}.
    Required by \texttt{action=terminate}.
    \item \textbf{\texttt{text}} \textit{(string).}
    Required by \texttt{action=answer}.
\end{evaluationitems}

\promptsection{Action: bash}
Run ONE shell command (requires \texttt{command}).
\begin{evaluationitems}
    \item CLI / file ops: write shell directly
    (\texttt{ls}, \texttt{cat}, \texttt{sed}, \texttt{grep},
    \texttt{python3 - <<'PY' ... PY}, \ldots).
    \item GUI: drive the screen with \texttt{pyautogui} inside a QUOTED
    heredoc, e.g.\end{evaluationitems}
\begin{evaluationcode}
\begin{verbatim}
python3 <<'PY'
import pyautogui
pyautogui.click(500, 300)  # coordinates are 0-999
                         # screen treated as 1000x1000
pyautogui.typewrite('hello', interval=0.02)
PY
\end{verbatim}
\end{evaluationcode}

\promptsection{PyAutoGUI Functions}
\begin{evaluationcode}
\begin{verbatim}
pyautogui.click(x, y)                       # left click
pyautogui.doubleClick(x, y)                 # double click
pyautogui.tripleClick(x, y)                 # triple click
pyautogui.rightClick(x, y)                  # right click
pyautogui.middleClick(x, y)                 # middle click
pyautogui.moveTo(x, y)                      # move mouse
pyautogui.dragTo(x, y, duration=0.5)        # drag to position
pyautogui.mouseDown()                       # press mouse button
pyautogui.mouseUp()                         # release mouse button
pyautogui.scroll(-5)                        # scroll down (negative=down)
pyautogui.press('enter')                    # press a key
pyautogui.hotkey('ctrl', 's')               # keyboard shortcut
pyautogui.typewrite('text', interval=0.02)  # type text
pyautogui.keyDown('shift')                  # hold key down
pyautogui.keyUp('shift')                    # release key
\end{verbatim}
\end{evaluationcode}
\begin{evaluationitems}
    \item Effects are seen via the NEXT screenshot; use \texttt{print()}
    for any text you need back (stdout is returned).
    \item One command may bundle multiple steps (several \texttt{pyautogui}
    lines in the heredoc; \texttt{\&\&} / pipes in shell).
    \item Optional \texttt{timeout} (seconds) for a slow command; the sandbox
    cuts commands short after approximately 30\,s, so split anything longer
    or run it in the background.
\end{evaluationitems}

\promptsection{Control Actions}
\begin{evaluationitems}
    \item \textbf{\texttt{wait}:} wait for the screen to settle.
    Optional \texttt{time} (seconds).
    \item \textbf{\texttt{terminate}:} finish the task.
    Requires \texttt{status = success | failure}.
    \item \textbf{\texttt{answer}:} answer a question-type task.
    Requires \texttt{text}.
\end{evaluationitems}

\promptsection{Function Call Format}
If you choose to call a function ONLY reply in the following format with NO suffix:
\begin{evaluationcode}
\begin{verbatim}
<tool_call>
<function=example_function_name>
<parameter=example_parameter_1>
value_1
</parameter>
</function>
</tool_call>
\end{verbatim}
\end{evaluationcode}

\promptsection{Important}
\texttt{<IMPORTANT>}
\begin{evaluationitems}
    \item Function calls MUST follow the specified format.
    \item The \texttt{action} parameter MUST be one of:
    \texttt{bash}, \texttt{wait}, \texttt{terminate}, \texttt{answer}.
    \item ALL GUI interactions MUST use \texttt{action=bash} with a
    \texttt{pyautogui} heredoc.
    \item Coordinates are 0--999 (the screen is a \texttt{1000x1000} grid).
    \item Heredoc delimiter MUST be quoted:
    \texttt{<<'PY'} (not \texttt{<<PY}).
    \item Observe effects via the NEXT screenshot; use \texttt{print()}
    for text output.
    \item When finished, use \texttt{action=terminate} (not bash exit commands).
    \item The current date is Thursday, September 24, 2026.
    \item Collapsed screenshots appear as:
    \texttt{This screenshot has been collapsed.}
\end{evaluationitems}
\texttt{</IMPORTANT>}

\promptsection{Response Format}
Every step:
\begin{list}{\arabic{enumi})}{%
    \usecounter{enumi}%
    \setlength{\leftmargin}{14pt}%
    \setlength{\itemsep}{3pt}%
    \setlength{\parsep}{0pt}%
    \setlength{\topsep}{2pt}}
    \item Action: one sentence describing your next move.
    \item A single \texttt{<tool\_call>...</tool\_call>} block.
\end{list}

\promptsection{Output Format Examples}
\textbf{GUI action (\texttt{action=bash} with pyautogui heredoc)}\par
Action: Click the ``File'' menu in the top menu bar.
\begin{evaluationcode}
\begin{verbatim}
<tool_call>
<function=computer_use>
<parameter=action>
bash
</parameter>
<parameter=command>
python3 <<'PY'
import pyautogui
pyautogui.click(50, 15)
PY
</parameter>
</function>
</tool_call>
\end{verbatim}
\end{evaluationcode}

\textbf{Shell command (\texttt{action=bash} with shell)}\par
Action: List files in the Documents folder.
\begin{evaluationcode}
\begin{verbatim}
<tool_call>
<function=computer_use>
<parameter=action>
bash
</parameter>
<parameter=command>
ls -la ~/Documents/
</parameter>
</function>
</tool_call>
\end{verbatim}
\end{evaluationcode}

\textbf{Wait (\texttt{action=wait})}\par
Action: Wait for the application to finish loading.
\begin{evaluationcode}
\begin{verbatim}
<tool_call>
<function=computer_use>
<parameter=action>
wait
</parameter>
<parameter=time>
3
</parameter>
</function>
</tool_call>
\end{verbatim}
\end{evaluationcode}

\textbf{Finish (\texttt{action=terminate})}\par
Action: The task is complete.
\begin{evaluationcode}
\begin{verbatim}
<tool_call>
<function=computer_use>
<parameter=action>
terminate
</parameter>
<parameter=status>
success
</parameter>
</function>
</tool_call>
\end{verbatim}
\end{evaluationcode}
\end{templatebox}

\subsection{Action-Schema Analysis: Separate GUI and CLI Tools}
\label{app:hybrid_system_prompt}

The following system prompt belongs to the \emph{Separate GUI/CLI tools}
variant in Table~\ref{tab:action_schema_analysis}
(Section~\ref{sec:action_schema_analysis}). Unlike the unified OSWorld
prompt above, it keeps the two interfaces in two distinct tools:
\texttt{computer\_use} for GUI interaction and \texttt{cli} for terminal
and file operations. HybridCUA instead exposes both through the single
\texttt{bash} action of Appendix~\ref{app:osworld_system_prompt}.

\begin{templatebox}{System Prompt: Separate GUI/CLI Tools}
\setlength{\parskip}{2pt}
\promptsection{Role}
You are a multi-purpose intelligent assistant. Based on my requests, you can
use tools to help me complete various tasks.
The password of the computer is \texttt{password}.

\promptsection{Tools}
You have access to the following functions:

\promptsection{GUI Tool: computer\_use}
\begin{evaluationcode}
\begin{verbatim}
Tool type:             function
Function name:         computer_use
Parameters type:       object
Required parameters:   ["action"]
\end{verbatim}
\end{evaluationcode}
Use a mouse and keyboard to interact with a computer, and take screenshots.
\begin{evaluationitems}
    \item This is an interface to a desktop GUI. For terminal commands and
    file operations, use the separate \texttt{cli} tool (it runs on the same
    machine); use this GUI for anything that must be seen or clicked on
    screen. You can also launch applications from the terminal via
    \texttt{cli} or by clicking desktop icons.
    \item Some applications may take time to start or process actions, so
    you may need to wait and take successive screenshots to see the results
    of your actions.
    \item The screen's resolution is \texttt{1000x1000}.
    \item Whenever you intend to move the cursor to click on an element
    like an icon, you should consult a screenshot to determine the
    coordinates of the element before moving the cursor.
    \item If you tried clicking on a program or link but it failed to load,
    even after waiting, try adjusting your cursor position so that the tip
    of the cursor visually falls on the element that you want to click.
    \item Make sure to click any buttons, links, icons, etc with the cursor
    tip in the center of the element. Don't click boxes on their edges
    unless asked.
\end{evaluationitems}

\promptsection{GUI Actions}
The \texttt{action} parameter is a required string with the following values:
\begin{evaluationitems}
    \item \textbf{\texttt{key}:} Performs key down presses on the arguments
    passed in order, then performs key releases in reverse order.
    \item \textbf{\texttt{type}:} Type a string of text on the keyboard.
    \item \textbf{\texttt{mouse\_move}:} Move the cursor to a specified
    \((x,y)\) pixel coordinate on the screen.
    \item \textbf{\texttt{left\_click}:} Click the left mouse button at a
    specified \((x,y)\) pixel coordinate on the screen. Optional
    \texttt{text} parameter can specify modifier keys (e.g.,
    \texttt{"ctrl"}, \texttt{"shift"}, \texttt{"ctrl+shift"}) that will
    be held during the click.
    \item \textbf{\texttt{left\_click\_drag}:} Click and drag the cursor to a
    specified \((x,y)\) coordinate.
    \item \textbf{\texttt{right\_click}:} Click the right mouse button at a
    specified \((x,y)\) pixel coordinate on the screen. Optional
    \texttt{text} parameter can specify modifier keys that will be held
    during the click.
    \item \textbf{\texttt{middle\_click}:} Click the middle mouse button at a
    specified \((x,y)\) pixel coordinate on the screen. Optional
    \texttt{text} parameter can specify modifier keys that will be held
    during the click.
    \item \textbf{\texttt{double\_click}:} Double-click the left mouse button
    at a specified \((x,y)\) pixel coordinate on the screen. Optional
    \texttt{text} parameter can specify modifier keys that will be held
    during the click.
    \item \textbf{\texttt{triple\_click}:} Triple-click the left mouse button
    at a specified \((x,y)\) pixel coordinate on the screen (simulated as
    double-click since it's the closest action). Optional \texttt{text}
    parameter can specify modifier keys that will be held during the click.
    \item \textbf{\texttt{scroll}:} Performs a scroll of the mouse scroll
    wheel. Optional \texttt{text} parameter can specify a modifier key
    (e.g., \texttt{"shift"}, \texttt{"ctrl"}) that will be held during
    scrolling.
    \item \textbf{\texttt{hscroll}:} Performs a horizontal scroll (mapped to
    regular scroll). Optional \texttt{text} parameter can specify a modifier
    key that will be held during scrolling.
    \item \textbf{\texttt{wait}:} Wait specified seconds for the change
    to happen.
    \item \textbf{\texttt{terminate}:} Terminate the current task and report
    its completion status.
    \item \textbf{\texttt{answer}:} Answer a question.
\end{evaluationitems}

\promptsection{GUI Tool Parameters}
\begin{evaluationitems}
    \item \textbf{\texttt{keys}} \textit{(array).}
    Required only by \texttt{action=key}.
    \item \textbf{\texttt{text}} \textit{(string).}
    Required by \texttt{action=type} and \texttt{action=answer}. Optional for
    click actions (\texttt{left\_click}, \texttt{right\_click},
    \texttt{middle\_click}, \texttt{double\_click}, \texttt{triple\_click})
    to specify modifier keys (e.g., \texttt{'ctrl'}, \texttt{'shift'},
    \texttt{'ctrl+shift'}). Optional for scroll actions (\texttt{scroll},
    \texttt{hscroll}) to specify a modifier key (e.g., \texttt{'shift'},
    \texttt{'ctrl'}) to hold during scrolling.
    \item \textbf{\texttt{coordinate}} \textit{(array).}
    \((x,y)\) coordinates.
    \item \textbf{\texttt{pixels}} \textit{(number).} Scroll amount.
    \item \textbf{\texttt{time}} \textit{(number).} Seconds to wait.
    \item \textbf{\texttt{status}} \textit{(string).}
    Task status for terminate. One of \texttt{success}, \texttt{failure}.
\end{evaluationitems}

\promptsection{Terminal Tool: cli}
\begin{evaluationcode}
\begin{verbatim}
Tool type:             function
Function name:         cli
Parameters type:       object
Required parameters:   ["action"]
Action type:           string
Action values:         ["bash"]
\end{verbatim}
\end{evaluationcode}
Run a shell command inside the machine (the same VM the GUI acts on).
The only supported \texttt{action} is \texttt{bash}:
\begin{evaluationitems}
    \item \textbf{\texttt{bash}:} run a shell command. Requires
    \texttt{command}; optional \texttt{timeout} (seconds).
\end{evaluationitems}
Output (stdout/stderr) is returned and also reflected in the next screenshot.

\promptsection{CLI Tool Parameters}
\begin{evaluationitems}
    \item \textbf{\texttt{command}} \textit{(string).}
    Required by \texttt{action=bash}: the shell command.
    \item \textbf{\texttt{timeout}} \textit{(number).}
    Optional for \texttt{action=bash}: timeout in seconds.
\end{evaluationitems}

\promptsection{Function Call Format}
If you choose to call a function ONLY reply in the following format with NO suffix:
\begin{evaluationcode}
\begin{verbatim}
<tool_call>
<function=example_function_name>
<parameter=example_parameter_1>
value_1
</parameter>
<parameter=example_parameter_2>
This is the value for the second parameter
that can span
multiple lines
</parameter>
</function>
</tool_call>
\end{verbatim}
\end{evaluationcode}

\promptsection{Important}
\texttt{<IMPORTANT>}\par
Reminder:
\begin{evaluationitems}
    \item Function calls MUST follow the specified format: an inner
    \texttt{<function=...></function>} block must be nested within
    \texttt{<tool\_call></tool\_call>} XML tags.
    \item Required parameters MUST be specified.
    \item You may provide optional reasoning for your function call in
    natural language BEFORE the function call, but NOT after.
    \item If there is no function call available, answer the question like
    normal with your current knowledge and do not tell the user about
    function calls.
    \item The current date is Thursday, September 24, 2026.
    \item Collapsed screenshots appear as text:
    \texttt{This screenshot has been collapsed.}
\end{evaluationitems}
\texttt{</IMPORTANT>}

\promptsection{Response Format}
Response format for every step:
\begin{list}{\arabic{enumi})}{%
    \usecounter{enumi}%
    \setlength{\leftmargin}{14pt}%
    \setlength{\itemsep}{3pt}%
    \setlength{\parsep}{0pt}%
    \setlength{\topsep}{2pt}}
    \item Action: a short imperative describing the next step --- a GUI
    interaction (\texttt{computer\_use}) or a terminal/file operation
    (\texttt{cli}).
    \item A single \texttt{<tool\_call>...</tool\_call>} block.
\end{list}
Rules:
\begin{evaluationitems}
    \item Output exactly in the order: Action, \texttt{<tool\_call>}.
    \item Be brief: one sentence for Action.
    \item Do not output anything else outside those parts.
    \item If finishing, use \texttt{action=terminate} in the tool call.
\end{evaluationitems}
\end{templatebox}

\subsection{Trajectory Construction: Domain CLI Skills}
\label{app:cli_skills}

The following eight domain-specific CLI skills provide practical guidance
for Qwen3.8-27B during trajectory collection in CUA-Gym, and are reproduced
here exactly as used. Their scope is limited to teacher sampling. They are not
included in the student model's supervised context, are not available during
RL rollouts, and appear in no evaluation prompt; the student therefore never
observes a skill at training or test time.

\begin{templatebox}{CLI Skill: GIMP}
\promptsection{GIMP --- CLI toolkit}
Two paths: \textbf{Pillow} for plain raster edits (resize, crop, rotate,
format convert, color tweaks) --- simplest and most reliable;
\textbf{GIMP batch Script-Fu} when the task needs a GIMP-specific feature
(layers, \texttt{.xcf}, filters/plugins, GIMP-exact output).

\promptsection{Pillow --- prefer this for ordinary image ops}
\begin{evaluationcode}
\begin{verbatim}
python3 <<'PY'
from PIL import Image, ImageEnhance, ImageFilter, ImageOps
im = Image.open("/home/user/photo.jpg")
print(im.size, im.mode)
im = im.resize((800, 600))
# or im.thumbnail((800,600)) to keep ratio
im = im.rotate(90, expand=True)
im = im.crop((left, top, right, bottom))
im = ImageOps.grayscale(im)
im = ImageEnhance.Brightness(im).enhance(1.2)
im = im.filter(ImageFilter.GaussianBlur(2))
im.convert("RGB").save("/home/user/out.png")
# convert("RGB") before saving jpg
PY
\end{verbatim}
\end{evaluationcode}

\promptsection{GIMP batch (Script-Fu) --- for .xcf / layers / GIMP filters}
The file usually is NOT open (GIMP is heavy); if it is:
\path{pkill -f gimp; sleep 3}. Run GIMP headless and quit at the end:
\begin{evaluationcode}
\begin{verbatim}
gimp -i -b '
  (let* ((image (car (gimp-file-load RUN-NONINTERACTIVE
                                   "/home/user/in.xcf" "in.xcf")))
         (drawable (car (gimp-image-flatten image))))
    (gimp-image-scale image 800 600)
    (file-png-save RUN-NONINTERACTIVE image drawable
                   "/home/user/out.png" "out" 0 9 1 1 1 1 1))
' -b '(gimp-quit 0)'
\end{verbatim}
\end{evaluationcode}
Handy Script-Fu calls: \texttt{gimp-image-flatten},
\texttt{gimp-image-scale}, \texttt{gimp-image-crop},
\texttt{gimp-item-transform-rotate-simple},
\texttt{gimp-image-convert-grayscale}, \texttt{gimp-text-fontname}
(add text), \texttt{file-jpeg-save} / \texttt{file-png-save} /
\texttt{gimp-xcf-save}.

\promptsection{Gotchas}
\begin{evaluationitems}
    \item Flatten (\texttt{gimp-image-flatten}) before exporting to a flat
    format like PNG/JPEG.
    \item \texttt{.xcf} output must use \texttt{gimp-xcf-save}, not
    \texttt{file-*-save}.
    \item Preserve the exact output size / format / path the instruction states.
    \item \texttt{gimp -i} (no UI) + always end with
    \path{-b '(gimp-quit 0)'} or the process hangs.
\end{evaluationitems}
\end{templatebox}

\begin{templatebox}{CLI Skill: LibreOffice Calc}
\promptsection{LibreOffice Calc --- CLI}
Edit on disk with openpyxl, re-read to verify, stop. Only the saved file
carries the document state, not an unsaved GUI buffer; \path{.~lock} is
advisory (doesn't block writes). Ensure lib:
\begin{evaluationcode}
\begin{verbatim}
python3 -c "import openpyxl" 2>/dev/null || pip install -q openpyxl
\end{verbatim}
\end{evaluationcode}
\textbf{Do NOT} pkill/reopen soffice, Ctrl+S in the GUI, or click a
``Document Recovery'' dialog --- it can restore the pre-edit version and
clobber your file.

\textbf{Write LITERAL values, not formula strings}
(\path{ws.cell(r,4).value = sales-cogs}) --- openpyxl does not evaluate
formulas, so \path{"=B2-C2"} is stored as a string with no cached value.
\begin{evaluationcode}
\begin{verbatim}
python3 <<'PY'
import openpyxl
p="/home/user/data.xlsx"; wb=openpyxl.load_workbook(p); ws=wb.active
print([ws.cell(1,c).value for c in range(1,ws.max_column+1)])
for r in range(2,ws.max_row+1):
    ws.cell(r,4).value = ws.cell(r,2).value - ws.cell(r,3).value  # literal
wb.save(p)
v=[openpyxl.load_workbook(p).active.cell(r,4).value
   for r in range(2,ws.max_row+1)]
assert all(isinstance(x,(int,float)) for x in v), v; print("OK",v)
PY
\end{verbatim}
\end{evaluationcode}
\begin{evaluationitems}
    \item Charts: \path{Reference(ws,min_col,min_row,max_col,max_row)} +
    \path{add_data(titles_from_data=True)} + \path{set_categories(cats)};
    not \texttt{chart\_type}/\path{Series.values=}. openpyxl quotes the
    sheet name in series references \(\rightarrow\) strip it via
    \path{series.val.numRef.f}/\path{strRef.f} for a bare reference.
    \item Pivot: when only the aggregated values are required, aggregate in
    Python to a static sheet (same-width header/data rows).
    \textbf{A native pivotTable object cannot be built by openpyxl; use
    UNO \texttt{DataPilotTables.createDataPilotDescriptor}.}
    Lookup/threshold: \texttt{bisect}, not an exact dict.
    \item Space in sheet name \(\rightarrow\) quote:
    \path{='Retail Price'.B2} (else Err:509). \texttt{0.00} not
    \texttt{0,00}. Colors need FF alpha (\texttt{FFFF0000}), exact hex.
    \texttt{.ods}\(\rightarrow\)odfpy,
    \texttt{.csv}\(\rightarrow\)pandas. Keep exact sheet/cell/OUTPUT path.
\end{evaluationitems}

\promptsection{Recalc / Calc-only}
Simplest:
\begin{evaluationcode}
\begin{verbatim}
soffice --headless --convert-to xlsx --outdir DIR FILE
\end{verbatim}
\end{evaluationcode}
(recomputes; never \texttt{--convert-to pdf}). Full engine via UNO: start
\begin{evaluationcode}
\begin{verbatim}
soffice --headless --accept="socket,host=localhost,port=2002;urp;" &
\end{verbatim}
\end{evaluationcode}
connect (\texttt{UnoUrlResolver}\(\rightarrow\)
\path{Desktop.loadComponentFromURL(...Hidden=True)}),
\path{doc.calculateAll(); doc.store()}.
\end{templatebox}

\begin{templatebox}{CLI Skill: LibreOffice Impress}
\promptsection{LibreOffice Impress --- CLI}
Edit the \texttt{.pptx} on disk with python-pptx, re-read to verify, stop.
Only the saved file carries the document state, not an unsaved GUI buffer;
\path{.~lock} is advisory. Ensure lib:
\begin{evaluationcode}
\begin{verbatim}
python3 -c "import pptx" 2>/dev/null || pip install -q python-pptx
\end{verbatim}
\end{evaluationcode}
\textbf{Do NOT} pkill/reopen soffice, Ctrl+S in the GUI, or click a
``Document Recovery'' dialog --- it can clobber your file. Prefer python-pptx
over clicking objects. Set formatting at the \textbf{run level} and read it
back (an object-level change may not reach run attributes).
\begin{evaluationcode}
\begin{verbatim}
python3 <<'PY'
from pptx import Presentation
from pptx.dml.color import RGBColor
p="/home/user/deck.pptx"; prs=Presentation(p); s=prs.slides[1]
def paint(tf):
    for para in tf.paragraphs:
        for r in para.runs:
            r.font.bold=True
            r.font.color.rgb=RGBColor(0xC9,0x21,0x1E)
for sh in s.shapes:
    if sh.has_text_frame: paint(sh.text_frame)
    if sh.has_table:
        for row in sh.table.rows:
            for c in row.cells: paint(c.text_frame)
            # don't skip table cells
prs.save(p)
cols={r.font.color.rgb for sh in Presentation(p).slides[1].shapes
      if sh.has_text_frame for para in sh.text_frame.paragraphs
      for r in para.runs}
assert RGBColor(0xC9,0x21,0x1E) in cols, cols; print("OK")
PY
\end{verbatim}
\end{evaluationcode}
\begin{evaluationitems}
    \item Notes: \path{s.notes_slide.notes_text_frame.text=...}.
    New slide: \path{prs.slides.add_slide(layout)}.
    Picture: \path{s.shapes.add_picture(img,left,top,width=Cm(8))};
    move via \path{shape.top/left}.
    \item Exact hex named (Dark Red 2 = \texttt{C9211E}, not
    \texttt{FF0000}); watch autocorrect curly \texttt{'}. \texttt{.odp}
    \(\rightarrow\)convert. Keep OUTPUT path.
    \item \path{font.name/size} stay \texttt{None} unless set explicitly on
    the run; \path{text=}/\texttt{add\_paragraph} drops the placeholder's
    inherited \texttt{pPr} (bullet \texttt{buChar}, size). Background,
    strikethrough, and bullet changes can fail without raising, so write
    them through the underlying \texttt{<a:p>} XML via lxml and read the
    result back.
\end{evaluationitems}

\promptsection{Export / Impress-only}
\begin{evaluationcode}
\begin{verbatim}
soffice --headless --convert-to pdf --outdir DIR FILE
\end{verbatim}
\end{evaluationcode}
or UNO: start
\begin{evaluationcode}
\begin{verbatim}
soffice --headless --accept="socket,host=localhost,port=2002;urp;" &
\end{verbatim}
\end{evaluationcode}
connect, then
\begin{evaluationcode}
\begin{verbatim}
doc.storeToURL("file:///.../deck.pdf", (
    PropertyValue(Name="FilterName",Value="impress_pdf_Export"),))
\end{verbatim}
\end{evaluationcode}
\end{templatebox}

\begin{templatebox}{CLI Skill: LibreOffice Writer}
\promptsection{LibreOffice Writer --- CLI}
Edit the \texttt{.docx} on disk with python-docx, re-read to verify, stop.
Only the saved file carries the document state, not an unsaved GUI buffer;
\path{.~lock} is advisory. Ensure lib:
\begin{evaluationcode}
\begin{verbatim}
python3 -c "import docx" 2>/dev/null || pip install -q python-docx
\end{verbatim}
\end{evaluationcode}
\textbf{Do NOT} pkill/reopen soffice, Ctrl+S in the GUI, or click a
``Document Recovery'' dialog --- it can clobber your file with the pre-edit
version.

Edit at the \textbf{run level} (\path{run.text}/\path{run.font}), never
\path{par.text=} (drops run formatting).
\begin{evaluationcode}
\begin{verbatim}
python3 <<'PY'
import docx
p="/home/user/report.docx"; d=docx.Document(p)
for par in d.paragraphs:
    for r in par.runs:
        if "OLD" in r.text: r.text=r.text.replace("OLD","NEW")
d.save(p)
d2=docx.Document(p)
assert not any("OLD" in r.text for par in d2.paragraphs for r in par.runs)
print("OK")
PY
\end{verbatim}
\end{evaluationcode}
\begin{evaluationitems}
    \item Run format: \path{run.bold=True}; \path{run.font.size=Pt(14)};
    \path{run.font.color.rgb=RGBColor(0xFF,0,0)}.
    \item Tables: \path{doc.tables}\(\rightarrow\)\path{table.rows}
    \(\rightarrow\)\path{cell.paragraphs}\(\rightarrow\)runs.
    \texttt{.odt}\(\rightarrow\)odfpy. Keep exact OUTPUT path.
    \item Export: \path{soffice --headless --convert-to pdf --outdir DIR FILE}.
    If the exported text drops soft-wrap trailing spaces, re-export via UNO
    with \path{FilterData=[UseTaggedPDF=True]}.
\end{evaluationitems}

\promptsection{Writer-only (update TOC/fields, mail merge)}
Start
\begin{evaluationcode}
\begin{verbatim}
soffice --headless --accept="socket,host=localhost,port=2002;urp;" &
\end{verbatim}
\end{evaluationcode}
connect (\texttt{UnoUrlResolver}\(\rightarrow\)
\path{Desktop.loadComponentFromURL(...Hidden=True)}), e.g.
\path{doc.getTextFields().refresh(); doc.store()}.
\end{templatebox}

\begin{templatebox}{CLI Skill: OS / Desktop}
\promptsection{OS / desktop --- CLI toolkit}
Filesystem, process, permission, package, and system-setting tasks are almost
entirely shell work --- the GUI file manager and settings dialogs are slower
and error-prone. Drive these from bash and verify by reading back state.

\promptsection{Files \& directories}
\begin{evaluationcode}
\begin{verbatim}
ls -la /home/user         # inspect permissions, sizes, hidden files
find /home/user -name '*.log' -mtime -1  # search by name/time
mkdir -p /home/user/a/b/c
cp -r src dst; mv old new; rm -rf junk
du -sh /home/user/*       # sizes; df -h for disk
\end{verbatim}
\end{evaluationcode}

\promptsection{Content edits \& search}
\begin{evaluationcode}
\begin{verbatim}
grep -rn "TODO" /home/user/proj
sed -i 's/old/new/g' /home/user/file.txt  # in-place edit
python3 - <<'PY'                        # structured/json edits
import json, pathlib
p = pathlib.Path("/home/user/config.json")
d = json.loads(p.read_text()); d["key"] = "value"
p.write_text(json.dumps(d, indent=2))
PY
\end{verbatim}
\end{evaluationcode}

\promptsection{Permissions \& ownership}
\begin{evaluationcode}
\begin{verbatim}
chmod 644 file; chmod +x script.sh; chmod -R 755 dir
sudo chown user:user file  # password is the sudo password in the prompt
\end{verbatim}
\end{evaluationcode}

\promptsection{Processes, archives, packages}
\begin{evaluationcode}
\begin{verbatim}
ps aux | grep -i firefox; pkill -f soffice.bin
tar -czf out.tar.gz dir/; tar -xzf in.tar.gz -C dest/; unzip a.zip -d dest/
sudo apt-get install -y <pkg>  # proxy is exported for the bash channel
\end{verbatim}
\end{evaluationcode}

\promptsection{GNOME desktop settings (wallpaper, theme, etc.)}
\begin{evaluationcode}
\begin{verbatim}
gsettings set org.gnome.desktop.background picture-uri \
    'file:///home/user/bg.png'
gsettings get org.gnome.desktop.interface gtk-theme
\end{verbatim}
\end{evaluationcode}

\promptsection{Gotchas}
\begin{evaluationitems}
    \item \texttt{sudo} needs \texttt{-S} with the password piped, or an
    interactive tty; the prompt states the password. Example:
    \path{echo "$PW" | sudo -S <cmd>}.
    \item \texttt{rm -rf} is irreversible --- double-check the path before
    running it.
    \item Verify every change (\texttt{ls -la}, \texttt{cat}, \texttt{stat},
    \texttt{gsettings get}) before terminating.
\end{evaluationitems}
\end{templatebox}

\begin{templatebox}{CLI Skill: PDF}
\promptsection{PDF --- CLI toolkit}
PDF tasks (stamp text, merge/split, rotate, extract text/images, count pages,
fill forms, redact) are almost always faster and more accurate via CLI than
GUI. PDFs have no persistent editing app holding a lock, so no
\texttt{pkill} dance is needed --- just read the input path and write the
exact OUTPUT path the instruction names.

\promptsection{Edit / annotate with PyMuPDF (fitz)}
\begin{evaluationcode}
\begin{verbatim}
python3 <<'PY'
import fitz
doc = fitz.open("/home/user/in.pdf")
print("pages:", doc.page_count)
for page in doc:
    print(page.get_text()[:200])  # extract text
    # (x,y) in points, origin top-left
    page.insert_text((72, 40), "Filed: April 1, 2026",
                     fontsize=10, fontname="times", color=(0, 0, 0))
    tw = fitz.get_text_length("Case No. 2026-CV-04521",
                             fontname="times", fontsize=10)
    page.insert_text((page.rect.width - 72 - tw, 40),
                     "Case No. 2026-CV-04521", fontsize=10,
                     fontname="times")
doc.save("/home/user/out.pdf")  # or doc.saveIncr() to edit in place
PY
\end{verbatim}
\end{evaluationcode}

\promptsection{Common recipes with pypdf}
\begin{evaluationcode}
\begin{verbatim}
python3 <<'PY'
from pypdf import PdfReader, PdfWriter
# merge
w = PdfWriter()
for f in ["/home/user/a.pdf", "/home/user/b.pdf"]:
    for pg in PdfReader(f).pages: w.add_page(pg)
with open("/home/user/merged.pdf", "wb") as fh: w.write(fh)

# split / rotate / extract a page range
r = PdfReader("/home/user/in.pdf")
w2 = PdfWriter()
for pg in r.pages[0:3]:
    pg.rotate(90)
    w2.add_page(pg)
with open("/home/user/first3_rotated.pdf", "wb") as fh: w2.write(fh)
PY
\end{verbatim}
\end{evaluationcode}

\promptsection{Other tools}
\begin{evaluationitems}
    \item \path{page.insert_image(rect, filename=...)} (fitz) to stamp a
    logo/image.
    \item \texttt{pdfplumber} for table extraction; \texttt{pikepdf} for
    encryption/metadata/repair.
    \item Make a PDF from a doc:
    \path{libreoffice --headless --convert-to pdf --outdir /home/user X.docx}.
    \item Rasterize a page to check visually:
    \path{page.get_pixmap(dpi=150).save("/tmp/p0.png")}.
\end{evaluationitems}

\promptsection{Gotchas}
\begin{evaluationitems}
    \item fitz coordinates are POINTS from the TOP-LEFT; y grows downward.
    \item Use exactly the font name/size the instruction specifies;
    \texttt{insert\_text} needs the y baseline, not the top of the glyph.
    \item Write to the precise OUTPUT path the instruction names, not back
    to the input path.
\end{evaluationitems}
\end{templatebox}

\begin{templatebox}{CLI Skill: VLC / Media}
\promptsection{VLC / media --- CLI toolkit}
Media tasks split cleanly: \textbf{inspect/convert/trim/extract}
\(\rightarrow\) \texttt{ffmpeg}/\texttt{ffprobe};
\textbf{edit tags/metadata} \(\rightarrow\) \texttt{mutagen};
\textbf{playback / playlist / snapshot} actions that must happen inside VLC
\(\rightarrow\) the \texttt{cvlc} command line or GUI. Only touch the VLC GUI
when the task is about VLC's own state (now-playing, playlist order, a setting).

\promptsection{Inspect with ffprobe}
\begin{evaluationcode}
\begin{verbatim}
ffprobe -v error -show_format -show_streams /home/user/clip.mp4  # full info
ffprobe -v error -show_entries format=duration -of csv=p=0 \
    /home/user/clip.mp4
\end{verbatim}
\end{evaluationcode}

\promptsection{Convert / trim / extract with ffmpeg}
\begin{evaluationcode}
\begin{verbatim}
ffmpeg -y -i in.mp4 -c:v libx264 -c:a aac out.mkv       # transcode
ffmpeg -y -ss 00:00:10 -to 00:00:25 -i in.mp4 \
    -c copy clip.mp4                                 # trim (fast copy)
ffmpeg -y -i in.mp4 -vn -acodec libmp3lame audio.mp3   # extract audio
ffmpeg -y -i in.mp4 -ss 5 -vframes 1 frame.png         # snapshot a frame
ffmpeg -y -i in.mp4 -vf scale=1280:720 out720.mp4       # resize
\end{verbatim}
\end{evaluationcode}

\promptsection{Read / write tags with mutagen}
\begin{evaluationcode}
\begin{verbatim}
python3 <<'PY'
from mutagen.easyid3 import EasyID3
from mutagen import File
print(File("/home/user/song.mp3").info.length)  # duration in seconds
a = EasyID3("/home/user/song.mp3")
a["title"] = "New Title"; a["artist"] = "Artist"; a.save()
PY
\end{verbatim}
\end{evaluationcode}

\promptsection{VLC command line (only when VLC itself must act)}
\begin{evaluationcode}
\begin{verbatim}
cvlc --play-and-exit /home/user/clip.mp4  # headless play
cvlc video.mp4 --video-filter=scene --scene-path=/tmp \
    --vout=dummy vlc://quit             # snapshots
\end{verbatim}
\end{evaluationcode}
VLC config/state lives at \path{~/.config/vlc/vlcrc}; recent-media in the
same dir.

\promptsection{Gotchas}
\begin{evaluationitems}
    \item \texttt{-c copy} only works when not re-encoding; drop it if you
    change codec/resolution.
    \item Always pass \texttt{-y} in batch so ffmpeg overwrites without
    prompting.
    \item Match the requested container/codec/bitrate and the exact output path.
\end{evaluationitems}
\end{templatebox}

\begin{templatebox}{CLI Skill: VS Code}
\promptsection{VS Code --- CLI toolkit}
Most VS Code tasks are really file/code tasks: create or edit source files,
run them, change settings, or manage extensions. Do the file work directly
in the shell/python --- it is exact and fast --- and use the \texttt{code}
CLI for editor-level state.

\promptsection{Edit files directly (don't hunt through the editor UI)}
\begin{evaluationcode}
\begin{verbatim}
cat > /home/user/proj/app.py <<'PY'
def add(a, b):
    return a + b
print(add(2, 3))
PY
python3 /home/user/proj/app.py  # run and check output
\end{verbatim}
\end{evaluationcode}
For surgical edits use python (read, replace, write) rather than retyping
in the editor.

\promptsection{Settings, keybindings, snippets (JSON on disk)}
\begin{evaluationcode}
\begin{verbatim}
~/.config/Code/User/settings.json  # e.g. {"editor.tabSize": 2}
~/.config/Code/User/keybindings.json
~/.config/Code/User/snippets/*.code-snippets
# workspace settings:
<project>/.vscode/settings.json
\end{verbatim}
\end{evaluationcode}
Edit these JSON files with python (\texttt{json.load} \(\rightarrow\)
mutate \(\rightarrow\) \texttt{json.dump}) to guarantee valid JSON.

\promptsection{The code CLI}
\begin{evaluationcode}
\begin{verbatim}
code --list-extensions          # what's installed
code --install-extension ms-python.python
code --uninstall-extension <id>
code -r /home/user/proj/app.py  # open a file in the running window
\end{verbatim}
\end{evaluationcode}
The on-disk file is the persistent state --- edit on disk and stop; don't
reopen/reload the editor just to make it visible (no GUI save needed).

\promptsection{Gotchas}
\begin{evaluationitems}
    \item Extension installs need network; the VM egress proxy is already
    exported for bash.
    \item Keep JSON valid --- a trailing comma breaks \texttt{settings.json}
    silently.
    \item Confirm the exact file path/name the task expects (case-sensitive).
\end{evaluationitems}
\end{templatebox}
\endgroup

\let\HybridCUAAppendixPreamble\undefined
\usepackage{iclr2027_conference,times}

\usepackage{amsmath,amsfonts,bm}

\def\eqref#1{equation~\ref{#1}}

\def\1{\bm{1}}

\DeclareMathAlphabet{\mathsfit}{\encodingdefault}{\sfdefault}{m}{sl}
\SetMathAlphabet{\mathsfit}{bold}{\encodingdefault}{\sfdefault}{bx}{n}

\usepackage{graphicx}
\usepackage{fontawesome5}
\usepackage{booktabs}
\usepackage[table]{xcolor}
\usepackage{wrapfig}
\usepackage{caption}
\captionsetup[table]{position=top}
\usepackage{hyperref}
\usepackage{url}

\title{HybridCUA: Learning to Orchestrate GUI and CLI for Computer-Use Agents}

\newif\ifHybridCUAPreprint
\HybridCUAPreprinttrue
\author{%
\begin{minipage}[t]{\dimexpr\textwidth-2\tabcolsep\relax}
\centering
Tongbo Chen\textsuperscript{1*}, Junbo Niu\textsuperscript{2*},
Zhengxi Lu\textsuperscript{1}, Niu Lian\textsuperscript{3},
Fei Tang\textsuperscript{1}, Yuchen Yan\textsuperscript{1}\\
Yike Hong\textsuperscript{1}, Yong Du\textsuperscript{1},
Yizhou Liu\textsuperscript{1}, Bofan Chen\textsuperscript{1},
Yongliang Shen\textsuperscript{1\textdagger}\\[0.5em]
{\normalfont\small
\textsuperscript{1}Zhejiang University \quad
\textsuperscript{2}Peking University \quad
\textsuperscript{3}Tsinghua University\\[0.3em]
\faGithub\ \href{https://github.com/ZJU-REAL/HybridCUA}{Code} \qquad
\faGlobe\ \href{https://zjureal.com/HybridCUA/}{Homepage} \qquad
\raisebox{-0.1em}{\includegraphics[height=1em]{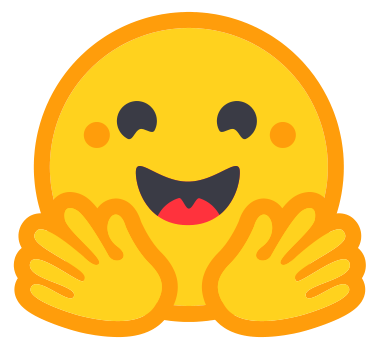}}\ \href{https://huggingface.co/collections/077lukamagic/hybridcua}{Hugging Face}}
\end{minipage}%
}
\ifHybridCUAPreprint
\iclrfinalcopy
\hypersetup{
  pdftitle={HybridCUA: Learning to Orchestrate GUI and CLI for Computer-Use Agents},
  pdfauthor={Tongbo Chen, Junbo Niu, Zhengxi Lu, Niu Lian, Fei Tang, Yuchen Yan, Yike Hong, Yong Du, Yizhou Liu, Bofan Chen, Yongliang Shen}
}
\fi

\begin{document}

\maketitle
\ifHybridCUAPreprint
\lhead{Preprint}
\begingroup
\renewcommand{\thefootnote}{\fnsymbol{footnote}}
\footnotetext[1]{Equal Contribution\qquad\textsuperscript{\textdagger} Corresponding Author}
\endgroup
\fi

\begin{abstract}

Computer use agents (CUAs) have demonstrated strong capabilities in completing
digital tasks. However, existing CUAs either rely solely on graphical user
interface (GUI) interactions, which are often inefficient and error prone, or augment
GUI interactions with application specific APIs or tools, which require
substantial engineering effort and are difficult to scale across applications.
We argue that the next generation of CUAs should combine GUI interactions with
the command line interface (CLI), leveraging the generality of the GUI and the
efficiency of shell commands. A critical challenge, however, is that current
models do not know when or how to use the CLI during task execution. To address
this challenge, we develop a data construction pipeline that produces three
types of trajectories: GUI only, CLI only, and interleaved GUI and CLI
trajectories. This pipeline results in \textbf{HybridCUA-8K}, containing 5K
hybrid trajectories and 3K verified RLVR tasks. Building on these data, we propose a
training framework with two stages: supervised fine tuning on the constructed trajectories,
followed by reinforcement learning with our CLI aware rewards that encourages agents to use the CLI selectively and reliably. Experiments show that \textbf{HybridCUA-9B} achieves
\textbf{53.6\%} accuracy on OSWorld, improving over the base model by
\textbf{14.8} percentage points, and improves performance on WindowsAgentArena
by \textbf{4.0} percentage points. These results demonstrate the effectiveness
and cross platform generalizability of the hybrid GUI and CLI paradigm for
computer use agents.

\end{abstract}

\begin{figure}[!h]
    \centering
    \includegraphics[width=0.84\linewidth]{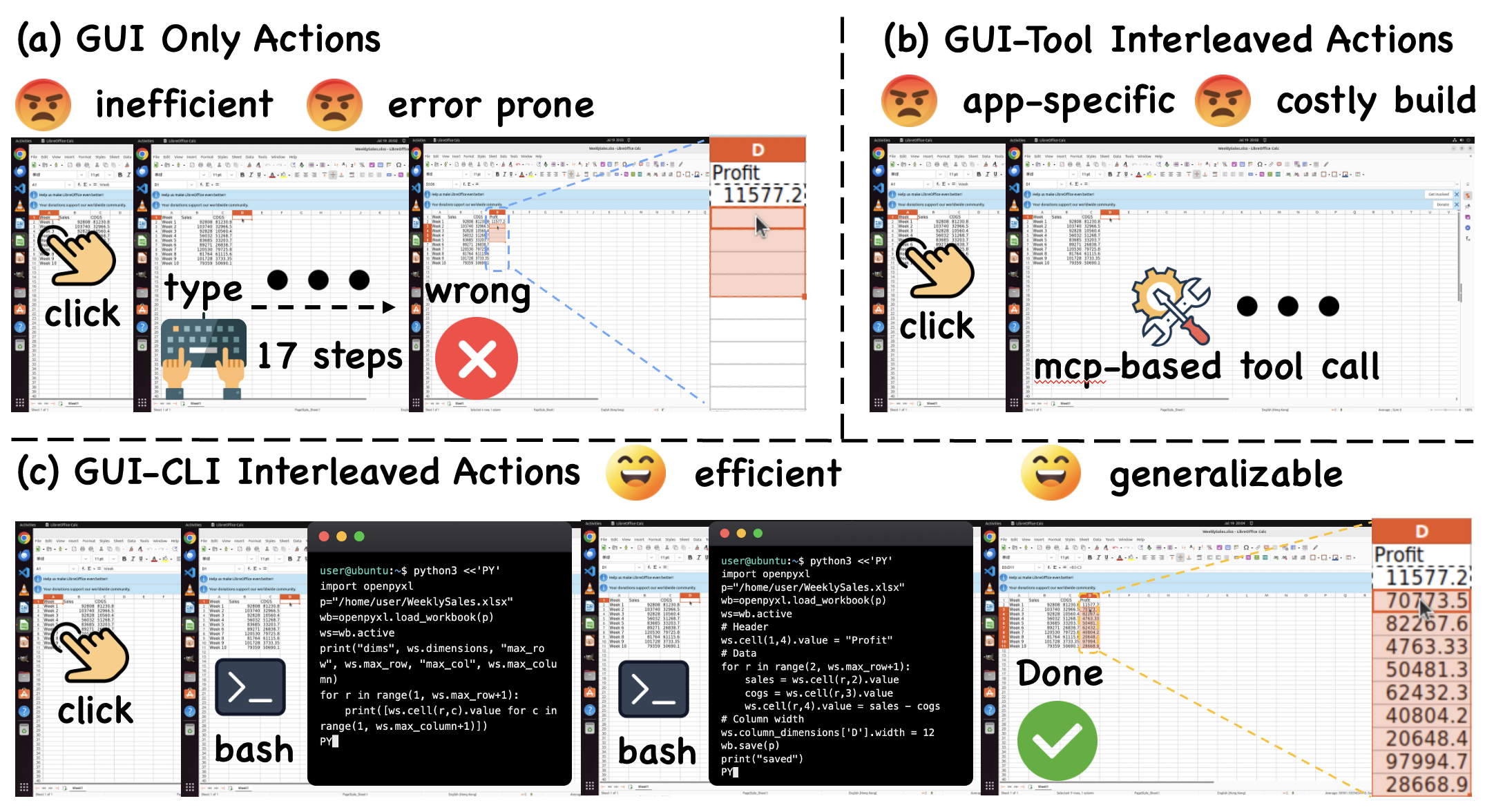}
    \caption{\textbf{Comparison of GUI only, GUI--tool, and GUI--CLI interleaved
    actions.} GUI only agents use low level clicks and keystrokes; GUI--tool
    agents invoke application specific tools; and GUI--CLI agents combine GUI
    interaction with general purpose shell commands.}
    \label{fig:gui_cli_overview}
\end{figure}

\section{Introduction}
\label{sec:introduction}

Computer-use agents (CUAs) built on multimodal large language models (MLLMs) typically act through clicks and keystrokes on graphical user interfaces (GUIs)---general, but slow and prone to cascading errors over long action sequences~\citep{qin2025uitars,wang2025opencua,xue2026evocua,lu2026uicopilot}. Application-specific APIs and tools speed things up, but only by sacrificing generality, as each must be built per application~\citep{jia2025osworldmcp,yang2025ultracua,hu2026toolcua}. Coding agents~\citep{openclaw2026peekaboo,claudecode2026,openai2026codex} suggest a way out: the command-line interface (CLI), which ships with the operating system and can collapse a long GUI sequence into a single command. We therefore argue that next-generation CUAs should be \emph{hybrid}, using the GUI for visual interaction and the shell for programmable operations (Figure~\ref{fig:gui_cli_overview}). Neither interface suffices on its own, but together they pair the generality of the GUI with the efficiency of code.

\begin{figure}[!b]
    \centering
    \includegraphics[width=0.96\linewidth]{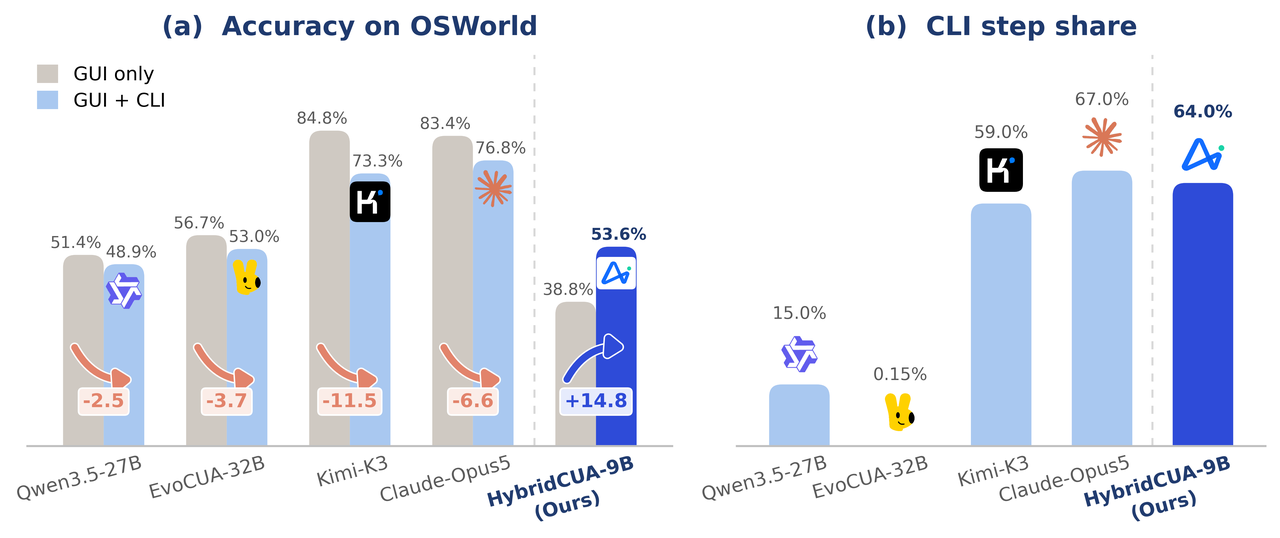}
    \caption{\textbf{The GUI--CLI orchestration problem.}
    \textbf{(a)} Adding CLI interface to existing agents reduces OSWorld
    accuracy, whereas the trained HybridCUA-9B improves over its GUI only
    counterpart. \textbf{(b)} CLI step share varies substantially across
    existing agents; HybridCUA learns to use the CLI selectively rather
    than merely maximizing command usage.}
    \label{fig:gui_cli_paradox}
\end{figure}

Hybrid interaction, however, does not come for free: simply handing a capable MLLM a shell does not help, but in fact hurts. As shown in Figure~\ref{fig:gui_cli_paradox}a, once the CLI is exposed, all four representative agents \emph{lose} 2.5 to 11.5 percentage points on OSWorld, indicating that they do not know \emph{how} to turn shell access into effective actions. Some do not even recognize \emph{when} to use it: Qwen3.5-27B~\citep{qwen2026qwen35} and EvoCUA-32B~\citep{xue2026evocua} route only 15.0\% and 0.15\% of their steps through the CLI (Figure~\ref{fig:gui_cli_paradox}b), grinding through long GUI sequences that a single command could replace. The bottleneck, in short, is not access to the shell but knowing when and how to use it.

We trace this deficit to two gaps in how current CUAs are trained. \textbf{First, the data never shows what hybrid behavior looks like.} GUI, shell, and code corpora are each abundant but siloed: CUA datasets contain almost exclusively GUI actions~\citep{wu2024osatlas}, whereas terminal and code datasets omit the GUI context in which commands are issued. \textbf{Second, the supervision is blind to interface choice.} Step-level imitation favors locally plausible actions, while outcome rewards merely check task completion~\citep{lai2025computerrl}; neither can tell a well-chosen interface from a successful but inefficient path.

We close both gaps with \textbf{HybridCUA}, a unified data and training framework for reliable GUI--CLI orchestration. Our key observation is that hybrid supervision need not be collected from scratch, since many GUI action sequences admit equivalent shell commands. Leveraging this, we construct a scalable pipeline spanning three execution schemas: GUI only trajectories converted from open-source datasets, CLI only trajectories sampled from strong models with an execution harness, and interleaved trajectories obtained by both free-form interface switching and GUI-to-CLI rewriting. The resulting 5{,}000 trajectories ground interface selection and command execution in GUI context. Beyond hybrid data construction, we further make interface choice an explicit training signal. Following supervised fine-tuning (SFT), we perform reinforcement learning with verifiable rewards (RLVR) in a live dual-interface environment, with two CLI-aware rewards mirroring the two missing abilities: a task-level reward $R_{\mathrm{CLI}}$ that teaches \emph{when} to invoke the CLI, and an action-level reward $R_{\mathrm{exec}}$ that teaches \emph{how}, using terminal feedback to improve command reliability.

The resulting \textbf{HybridCUA-9B} reaches 53.6\% on OSWorld, 14.8 percentage points above its base model and the best among comparably sized models. Notably, it uses the CLI for 64.0\% of its steps, on par with the most CLI-heavy baseline, yet \emph{gains} where that baseline loses (Figure~\ref{fig:gui_cli_paradox}), confirming that the improvement comes from knowing when and how to use the shell rather than from using it more. The gains carry over to Windows: on WindowsAgentArena, HybridCUA-9B reaches 36.0\%, a 4.0-point improvement over the base model.

Our main contributions are summarized as follows:

\begin{itemize}
    \item We develop a scalable data generation pipeline and construct \textbf{HybridCUA-8K}, comprising 5{,}000 GUI only, CLI only, and interleaved trajectories together with 3{,}000 verified RLVR tasks, each labeled by whether CLI use offers a clear execution advantage.
    \item We propose a two-stage training framework that performs SFT on the 5{,}000 trajectories, followed by RLVR on the 3{,}000 verified tasks with CLI-aware rewards that separately target routing and execution errors.
    \item We train \textbf{HybridCUA-9B}, which achieves 53.6\% on OSWorld, improving over the base model by 14.8 percentage points. We will release the trajectories, RLVR tasks, data generation and training pipelines, and the HybridCUA-9B model.
\end{itemize}

\section{Related Work}
\label{sec:related_work}

\subsection{GUI-Based Computer-Use Agents}

Recent computer-use agents leverage large scale GUI dataset, and interaction trajectories to jointly improve visual
understanding, visual grounding, task planning, and reasoning
\citep{wu2024osatlas,lu2026uir1,qin2025uitars,wang2025opencua,lu2025uis1}. These
advances enable a single policy to operate across diverse applications and
platforms through screenshots and human-like mouse and keyboard actions.
More recently, CUA-Gym and ScaleCUA scale reinforcement learning with
verifiable rewards through automatic task generation, scalable environments,
and efficient online training \citep{wang2026cuagym,lv2026scalecua}. However,
these agents still operate primarily through atomic actions such as clicking,
typing, and scrolling. Tasks involving file manipulation, text editing, or
repeated operations may therefore require long GUI sequences even when concise
programmatic solutions exist, reducing efficiency and increasing the risk of
cascading errors.

\subsection{Hybrid Computer Use}

Recent work extends computer-use agents beyond GUI only interaction. MCPWorld
and OSWorld-MCP benchmark coordination between GUI actions and APIs or MCP
tools, while ComputerRL, UltraCUA, ToolCUA, and UI-TARS-2 learn policies over
related hybrid action spaces \citep{yan2025mcpworld,jia2025osworldmcp,
lai2025computerrl,yang2025ultracua,hu2026toolcua,wang2025uitars2}. Such tools
can compress repetitive low level interactions, but their coverage and
portability are bounded by predefined, application-specific interfaces.

The shell escapes this constraint, and recent work has begun to explore
GUI--CLI workflows through benchmarks, environments, and agent systems
\citep{li2026weavebench,yang2026clianything,zhou2026qwen_ui_agent}. The two
closest efforts differ mainly in how they supervise hybrid behavior.
CUA-Universe synthesizes hybrid tasks at scale and collects guided
trajectories, but scores them with a VLM rather than task-specific
programmatic verification, leaving feedback too coarse for online training
\citep{shi2026cuauniverse}. RecreationWorld does offer verifiable
application-recreation environments, yet supervises task outcomes alone, so
interface selection remains implicit in trajectory imitation
\citep{bai2026recreationworld}. In neither case is the \emph{choice} of
interface part of the learning signal. HybridCUA closes this gap by pairing
executable task verifiers with CLI-aware rewards that directly supervise
interface routing and command reliability.

\section{HybridCUA}
\label{sec:method}

\subsection{Hybrid Computer-Use Formulation}
\label{sec:formulation}
\paragraph{Environment and observation.}
We formulate computer use as a partially observable Markov decision process
$\mathcal{M}=\langle\mathcal{S},\mathcal{A},P,R,\Omega,O,\gamma\rangle$.
The latent state $s_t\in\mathcal{S}$ contains the complete computer state,
whereas the agent observes only
\begin{equation}
    o_t=(I_t,\tilde{y}_{t-1})\sim O(\cdot\mid s_t),
    \label{eq:observation}
\end{equation}
where $I_t$ is the current screenshot and $\tilde{y}_{t-1}$ is the
stdout/stderr of the preceding CLI action, or empty otherwise. Each action
returns one post action screenshot, paired with its CLI output when available.

\paragraph{Unified GUI--CLI action space.}
Rather than exposing separate GUI and CLI, all executable interactions
use the same \texttt{bash} action. We define the unified executable action
space as
\begin{equation}
    \mathcal{A}
    =
    \left\{
        \mathrm{bash}(c,\delta)
        \;\middle|\;
        c\in\mathcal{C}_{\mathrm{CLI}}\cup\mathcal{C}_{\mathrm{GUI}}
    \right\}
    \cup
    \left\{
        \mathrm{wait},\,
        \mathrm{terminate},\,
        \mathrm{answer}
    \right\},
    \label{eq:action_encoding}
\end{equation}
where $c$ is the command and $\delta$ is an optional timeout.
$\mathcal{C}_{\mathrm{CLI}}$ contains direct shell commands and
$\mathcal{C}_{\mathrm{GUI}}$ contains quoted Python heredocs with one or more
\texttt{pyautogui} code. The complete action schema
is provided in Appendix Table~\ref{tab:action_space}.

\subsection{Scalable Hybrid Data Generation Pipeline}
\label{sec:trajectory_construction}

\begin{figure}[t]
      \centering
      \includegraphics[width=\linewidth]{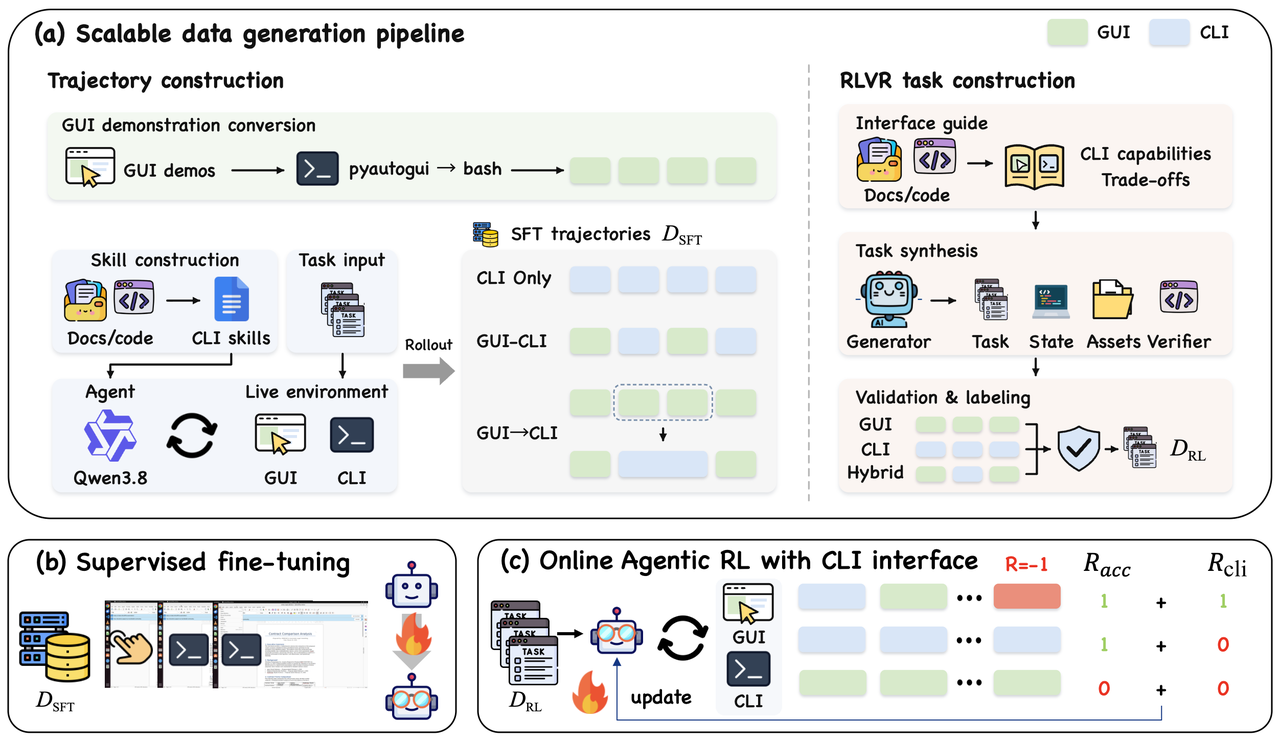}
      \caption{\textbf{The HybridCUA data and training pipeline.}
      \textbf{(a)} Scalable generation of GUI only, CLI only, and interleaved
      GUI--CLI trajectories, together with annotated RL tasks.
      \textbf{(b)} Supervised fine tuning on
      $\mathcal{D}_{\mathrm{SFT}}$.
      \textbf{(c)} Online agentic reinforcement learning with CLI aware reward
      signals.}
      \label{fig:method_overview}
  \end{figure}

Accordingly, our scalable pipeline in
Figure~\ref{fig:method_overview}(a) constructs SFT trajectories across all
three interface modes and synthesizes RLVR tasks labeled by whether CLI use
offers a clear execution advantage.

\paragraph{Trajectory construction across interface modes.}
\textbf{GUI only.}
We convert open source UI-MOPD trajectories~\citep{lian2026uimopd},
re-expressing each GUI action as equivalent \texttt{pyautogui} code in our
unified action format.
\textbf{CLI only.}
For each application, we distill its documentation and reference code into
an application-specific CLI skill listing the available CLI interfaces,
common commands, and representative usage patterns, used only for this
collection. Equipped with these skills, a Claude Code harness, and the CLI
interface alone, Qwen3.8-27B~\citep{qwen2026qwen38} solves tasks sampled from
CUA-Gym~\citep{wang2026cuagym}, and we retain its successful rollouts.
\textbf{Interleaved GUI--CLI.}
Two complementary routes yield trajectories spanning both interfaces. In
the first, Qwen3.8-27B accesses both and chooses between them during
execution. In the second, we rewrite suitable segments of its GUI only
trajectories into equivalent CLI commands, then replay the result and keep
only successful replays. Appendix~\ref{app:sft_data} details all three routes.

\paragraph{RLVR task construction and CLI advantage labeling.}
Building on the CUA-Gym pipeline, we construct 3{,}000 verified tasks and label
whether CLI use provides a clear execution advantage. For each application, an
interface guide built from its documentation and reference code lists the
available CLI interfaces, the operations where CLI is more efficient or
reliable, and those that require or favor the GUI, letting the task generator
locate the capability boundary between the two interfaces and compose hybrid
tasks. The pipeline then synthesizes the task, its environment states, assets,
and verifier, keeping those with an executable verifier and a reachable
target state. To label a retained task, we sample 16 rollouts under each of
GUI only, CLI only, and GUI--CLI and rank the three modes by success rate,
breaking near-ties by step count. We assign $b^\star=1$ to tasks where CLI
use offers a clear execution advantage, and $b^\star=0$ otherwise. All tasks
are retained.
Appendix~\ref{app:task_labels} reports thresholds and label counts.

\subsection{Two Stage Training Paradigm}
\label{sec:training}

We train HybridCUA in two stages (Figure~\ref{fig:method_overview}(b), (c)).
Supervised fine-tuning teaches GUI--CLI execution in the unified action format
and interface switching, while reinforcement learning with verifiable rewards
improves interface selection and CLI execution through online interaction.

\paragraph{Stage I: supervised warm-up.}
We combine the three trajectory types into the supervised training set
$\mathcal{D}_{\mathrm{SFT}}$. Each trajectory is
serialized as a task instruction followed by observations and
actions in the unified action format. We optimize the next token prediction
loss over agent generated responses:
\begin{equation}
    \mathcal{L}_{\mathrm{SFT}}
    =
    -\mathbb{E}_{\tau\sim\mathcal{D}_{\mathrm{SFT}}}
    \left[
        \sum_{t=0}^{T}
        \log \pi_\theta(a_t\mid x,h_t)
    \right],
    \label{eq:sft_objective}
\end{equation}
where $x$ is the instruction and
$h_t=(o_0,a_0,\ldots,o_t)$ is the interaction history. The three trajectory
types provide complementary supervision for GUI control, shell command, and interface routing.

\paragraph{Stage II: CLI aware online agentic reinforcement learning.}
We next optimize the model through online interaction with a live GUI-CLI
environment. Standard agentic RL typically uses a final accuracy reward $R_{\mathrm{acc}}$, which
encourage task completion but provide limited supervision
for interface selection and CLI reliability. Successful rollouts receive the
same accuracy reward despite unnecessary interface switches, while failed CLI
commands go unpenalized if the task is eventually completed. This weak credit
assignment makes it difficult to learn when and how to use the CLI.

To address these issues, we introduce a CLI aware reward consisting of a
task level CLI use signal and a step level command execution signal. Each of
the 3{,}000 verified tasks has the binary annotation $b^\star$ defined above.
For a rollout $\tau$, let
$b(\tau)=\mathbb{I}[\tau\text{ contains at least one direct CLI command}]$.
We define the CLI reward as
\begin{equation}
    R_{\mathrm{CLI}}
    =
    \mathbb{I}\!\left[\mathrm{Success}(\tau)\right]
    \mathbb{I}\!\left[b(\tau)=b^\star\right].
    \label{eq:cli_reward}
\end{equation}
By rewarding successful rollouts whose CLI usage matches the task-specific
label, $R_{\mathrm{CLI}}$ encourages the model to select the CLI when it offers
a clear execution advantage and avoid unnecessary CLI use otherwise. It
thereby teaches task dependent interface selection. For command execution,
terminal feedback provides localized supervision. We define the step-level
execution reward as
\begin{equation}
    r_t^{\mathrm{exec}} =
    \begin{cases}
        -1, & \text{if the command incurs a shell-level execution failure},\\
        0,  & \text{otherwise},
    \end{cases}
    \label{eq:exec_reward}
\end{equation}
and set $r_t^{\mathrm{exec}}=0$ at non-CLI steps. The two CLI-aware signals
operate at complementary granularities. Because $R_{\mathrm{CLI}}$ evaluates
whether the rollout selects an appropriate interface strategy, we incorporate
it into the trajectory level reward:
\begin{equation}
    R(\tau)
    =
    R_{\mathrm{acc}}
    + \lambda_{\mathrm{CLI}}R_{\mathrm{CLI}}.
    \label{eq:total_reward}
\end{equation}
In contrast, $r_t^{\mathrm{exec}}$ 
attribute a shell execution failure to the responsible CLI action. For a
GRPO group of $G$ rollouts, let $\mu_G$ and $\sigma_G$ denote the mean and
standard deviation of their trajectory rewards. We augment the normalized
trajectory advantage only for tokens belonging to action $a_t$:
\begin{equation}
    \widehat{A}_{t,j}
    =
    \frac{R(\tau)-\mu_G}{\sigma_G+\epsilon}
    + \lambda_{\mathrm{exec}}r_t^{\mathrm{exec}},
    \qquad j\in\operatorname{Tok}(a_t),
    \label{eq:step_advantage}
\end{equation}
where $\epsilon$ ensures numerical stability.

\section{Experiments}
\label{sec:experiments}

\subsection{Experimental Settings}
\label{sec:experimental_setup}

\paragraph{Implementation details.}
We initialize HybridCUA-9B from Qwen3.5-9B~\citep{qwen2026qwen35} and adopt a
two-stage training pipeline: supervised warm-up for 2 epochs on the three
trajectory types, followed by online agentic RL with GRPO using our
constructed RLVR tasks, where the CLI aware reward uses
$\lambda_{\mathrm{CLI}}=0.1$ at the trajectory level and
$\lambda_{\mathrm{exec}}=0.3$ at the step level.
We use \texttt{verl}~\citep{sheng2025hybridflow} and
Megatron-LM~\citep{shoeybi2019megatron} for supervised training, and
\texttt{slime}~\citep{thudm2025slime} with Megatron-LM and
SGLang~\citep{zheng2024sglang} for RL optimization and rollout. Dataset statistics and full training
configurations are provided in Appendices~\ref{app:dataset_details}
and~\ref{app:training_details}.

\paragraph{Baselines and benchmarks.}
We use OSWorld~\citep{xie2024osworld} as our primary benchmark and compare
HybridCUA-9B against three categories of baselines:
(i) GUI only agents, including Qwen3.5~\citep{qwen2026qwen35},
OpenCUA~\citep{wang2025opencua}, and EvoCUA~\citep{xue2026evocua};
(ii) GUI--API agents, including ToolCUA~\citep{hu2026toolcua},
AutoGLM-OS-9B~\citep{lai2025computerrl}, and
UltraCUA~\citep{yang2025ultracua}; and
(iii) Qwen3.5-9B equipped with either the GUI only or GUI--CLI action space.
Following the official evaluation protocol, we report task accuracy
(\textbf{Acc.}) and average steps (\textbf{Avg. Steps}), with a
maximum 50 steps per task. To evaluate out-of-distribution generalization
along two complementary axes, we further test HybridCUA-9B on OSWorld-MCP and
WindowsAgentArena~\citep{bonatti2024windowsagentarena}. OSWorld-MCP measures
transfer to an MCP-enabled environment, whereas WindowsAgentArena measures
cross operating-system transfer.

\subsection{Main Results}
\label{sec:main_results}

\begin{table*}[t]
    \centering
    \small
    \setlength{\tabcolsep}{8pt}
    \caption{Main results on OSWorld. ``Action space'' denotes the executable
    interfaces available to each model. Accuracy is reported in percentage.
    Lower average steps indicate shorter trajectories.
    }
    \label{tab:main_results}
    \begin{tabular}{lccc}
        \toprule
        \textbf{Model} & \textbf{Action space} & \textbf{Acc.$\uparrow$} & \textbf{Avg.} \textbf{Steps$\downarrow$} \\
        \midrule
        \rowcolor{gray!10}
\multicolumn{4}{l}{
    \textit{General Models}
} \\
        Qwen3.5-27B~\citep{qwen2026qwen35} & GUI & 51.4 & 25.7 \\
        Qwen3.5-35B-A3B~\citep{qwen2026qwen35} & GUI & 50.4 & 26.3 \\
        \midrule
        \rowcolor{gray!10}
\multicolumn{4}{l}{
    \textit{Specialized CUA Models}
} \\
        OpenCUA-7B~\citep{wang2025opencua} & GUI & 28.1 & 19.8 \\
        OpenCUA-32B~\citep{wang2025opencua} & GUI & 34.1 & 18.7 \\
        EvoCUA-8B~\citep{xue2026evocua} & GUI & 46.1 & 31.1 \\
        EvoCUA-32B~\citep{xue2026evocua} & GUI & 56.7 & 25.0 \\
        ToolCUA-8B~\citep{hu2026toolcua} & GUI+API & 46.8 & 14.9 \\
        AutoGLM-OS-9B~\citep{lai2025computerrl} & GUI+API & 48.9 & -- \\
        UltraCUA-32B~\citep{yang2025ultracua} & GUI+API & 43.7 & -- \\
        \midrule
        \rowcolor{gray!10}
\multicolumn{4}{l}{
    \textit{Ours}
} \\
        Qwen3.5-9B~\citep{qwen2026qwen35} & GUI & 38.8 & 31.6 \\
        \quad + SFT & GUI
        & 44.2
          {\scriptsize\textcolor{green!35!black}{\textbf{$+5.4$}}}
        & 26.3
          {\scriptsize\textcolor{green!35!black}{\textbf{$-5.3$}}} \\
        \quad + SFT + RL & GUI
        & 50.4
          {\scriptsize\textcolor{green!35!black}{\textbf{$+11.6$}}}
        & 22.1
          {\scriptsize\textcolor{green!35!black}{\textbf{$-9.5$}}} \\
        \cmidrule{1-4}
        Qwen3.5-9B~\citep{qwen2026qwen35} & GUI+CLI
        & 18.4
          {\scriptsize\textcolor{red!55!black}{\textbf{$-20.4$}}}
        & 22.1
          {\scriptsize\textcolor{green!35!black}{\textbf{$-9.5$}}} \\
        \quad + SFT & GUI+CLI
        & 46.0
          {\scriptsize\textcolor{green!35!black}{\textbf{$+7.2$}}}
        & 19.8
          {\scriptsize\textcolor{green!35!black}{\textbf{$-11.8$}}} \\
        \rowcolor[RGB]{232,240,251}
        \quad \textbf{HybridCUA-9B} & \textbf{GUI+CLI}
        & \textbf{53.6}
          {\scriptsize\textcolor{green!35!black}{\textbf{$+14.8$}}}
        & \textbf{14.0}
          {\scriptsize\textcolor{green!35!black}{\textbf{$-17.6$}}} \\
        \bottomrule
    \end{tabular}
\end{table*}

\textbf{GUI--CLI vs. GUI only.} In Table~\ref{tab:main_results}, HybridCUA-9B
reaches \textbf{53.6\%} accuracy with \textbf{14.0} average steps, the best on
both metrics among comparably sized models. Merely exposing the CLI is instead
harmful: steps fall by 9.5 but accuracy drops from 38.8\% to 18.4\%, so shorter
trajectories here reflect early failure, not efficiency. Once trained to use
it, the second interface pays off at both stages, 46.0\% versus 44.2\% after
SFT and 53.6\% versus 50.4\% after RL, with steps cut from 26.3 to 19.8 and
from 22.1 to 14.0. Both branches share the same base model, comparable SFT
corpora, and identical RLVR tasks and RL steps, isolating the added interface.

\textbf{GUI--CLI vs. GUI--API.} HybridCUA-9B exceeds AutoGLM-OS-9B by 4.7
percentage points, ToolCUA-8B by 6.8, and the four times larger UltraCUA-32B by
9.9, with 0.9 fewer steps than ToolCUA-8B. All three rely on
application-specific APIs, whereas the shell requires no per-application
construction.

\subsection{Ablation Study}
\label{sec:training_ablation}

We ablate the two training stages in turn: first the composition of the
supervised corpus, then the contribution of online RL and its CLI-aware
reward.

\paragraph{Effect of SFT data composition.}
\setlength{\columnsep}{8pt}
\begin{wraptable}{r}{0.46\linewidth}
    \centering
    \small
    \caption{SFT data ablation on OSWorld.}
    \label{tab:sft_data_ablation}
    \setlength{\tabcolsep}{4pt}
    \renewcommand{\arraystretch}{1.05}
    \begin{tabular}{lcc}
        \hline
        Configuration & Acc.$\uparrow$ & Avg. Steps$\downarrow$ \\
        \hline
        Qwen3.5-9B & 38.8 & 31.6 \\
        \hspace{0.6em}w/ CLI only & 31.7 & 19.1 \\
        \hspace{0.6em}w/ GUI only & 43.2 & 29.5 \\
        \hspace{0.6em}w/ Hybrid only & 41.0 & 22.6 \\
        \rowcolor[RGB]{220,232,247}
        \textbf{HybridCUA-9B-SFT} & \textbf{46.0} & \textbf{19.8} \\
        \hline
    \end{tabular}
\end{wraptable}

Table~\ref{tab:sft_data_ablation} shows that \textbf{mixing all three
trajectory types yields higher accuracy than any single type SFT}.
The mixed corpus achieves 46.0\% accuracy, outperforming GUI only,
CLI only, and hybrid only SFT by 2.8, 14.3, and 5.0 percentage points,
respectively. It averages 19.8 steps, only 0.7 more than CLI only SFT.
Notably, interleaved trajectories alone do not match the mixed corpus,
suggesting that single interface demonstrations and hybrid trajectories
provide complementary supervision.

\paragraph{Effect of RL.}
Starting from the mixed-data SFT checkpoint, RL increases accuracy from
46.0\% to \textbf{53.6\%} and reduces average steps from 19.8 to
\textbf{14.0} (Tables~\ref{tab:sft_data_ablation}
and~\ref{tab:main_results}). This yields a \textbf{7.6} percentage-point
accuracy gain and saves \textbf{5.8} steps per task on average, a
\textbf{29.3\%} reduction relative to SFT. Thus, RL improves task completion
while further shortening trajectories beyond supervised warm-up.

\begin{figure}[!htb]
    \centering
    \includegraphics[width=\linewidth]{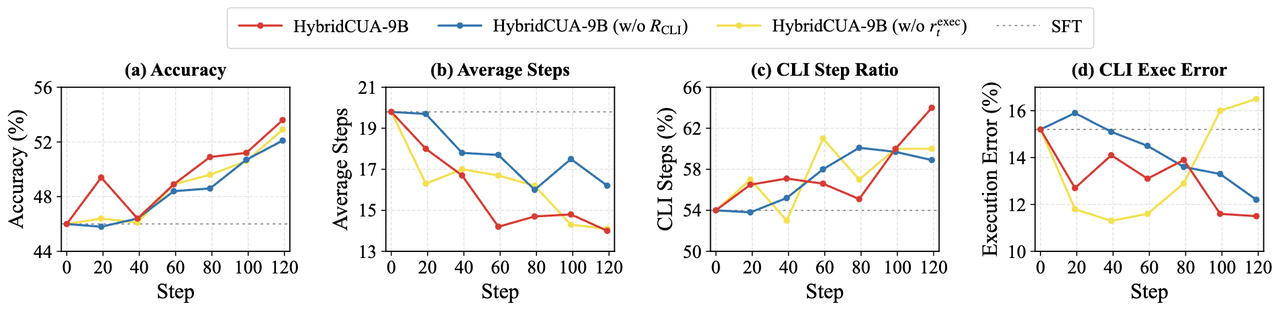}
    \caption{RL ablation curves on OSWorld. The four panels report task
    accuracy, average environment steps, the percentage of executable steps
    issued through the CLI, and CLI execution error rate.}
    \label{fig:training_ablation}
\end{figure}

To isolate the contribution of each CLI-aware signal, we compare three RL
variants in Figure~\ref{fig:training_ablation}, all initialized from the same
SFT checkpoint under an identical GRPO budget: the full HybridCUA-9B, a
variant without the task-level $R_{\mathrm{CLI}}$, and a variant without the
step-level $r_t^{\mathrm{exec}}$, with the SFT checkpoint shown as a
reference line. The two signals govern different quantities. Removing
$R_{\mathrm{CLI}}$ costs little accuracy but stalls the
efficiency gain: CLI usage reaches only 58.9\% instead of 64.0\%, and
trajectories shorten by 18.2\% rather than 29.3\% relative to SFT. Removing
$r_t^{\mathrm{exec}}$ leaves accuracy and steps nearly intact, yet execution errors climb back to 16.5\% by step 120, above the SFT
level, while the full model settles at 11.5\%. The task-level reward
therefore decides \emph{when} to use the CLI, whereas the step-level
reward governs \emph{how reliably} commands execute.

\subsection{Out-of-Distribution Generalization}
\label{sec:ood_generalization}

\begingroup
\begin{table}[!ht]
    \centering
    \caption{Out-of-distribution generalization on OSWorld-MCP and
    WindowsAgentArena.}
    \label{tab:ood_results}
    \small
    \setlength{\tabcolsep}{4pt}
    \renewcommand{\arraystretch}{1.08}
    \begin{tabular}{@{}lccc@{}}
        \toprule
        Model & Action space & OSWorld-MCP & WindowsAgentArena \\
        \midrule
        Qwen3-VL-8B-Instruct~\citep{bai2025qwen3vl} & GUI & 28.2 & 26.4 \\
        Qwen3-VL-235B-A22B~\citep{bai2025qwen3vl} & GUI & 38.1 & 32.1 \\
        Qwen3.5-9B~\citep{qwen2026qwen35} & GUI & 38.0 & 32.0 \\
        ToolCUA-8B~\citep{hu2026toolcua} & GUI + API & 46.8 & 33.8 \\
        \rowcolor[RGB]{220,232,247}
        \textbf{HybridCUA-9B (ours)} & \textbf{GUI + CLI}
        & \textbf{47.1} & \textbf{36.0} \\
        \bottomrule
    \end{tabular}
\end{table}
\endgroup

As shown in Table~\ref{tab:ood_results}, HybridCUA-9B achieves
\textbf{47.1\%} accuracy on OSWorld-MCP, exceeding Qwen3.5-9B by
\textbf{9.1} percentage points and performing comparably to ToolCUA-8B. This suggests that learned GUI--CLI orchestration transfers to
an MCP-enabled environment while remaining competitive with
application-specific GUI--API integration. For cross operating-system
transfer, HybridCUA-9B reaches \textbf{36.0\%} on WindowsAgentArena,
outperforming Qwen3.5-9B and ToolCUA-8B by \textbf{4.0}
and \textbf{2.2} percentage points, respectively. Notably, it issues
PowerShell commands despite training only on Linux shells, suggesting that
what transfers is \emph{when} to delegate to the shell rather than memorized
commands. Together, these gains
across both environments support our motivation to learn \emph{when} and
\emph{how} to combine GUI and CLI as a transferable alternative to
application-specific tool integration.

\subsection{Analysis}
\label{sec:analysis}

\begin{figure}[!htb]
    \centering
    \includegraphics[width=\linewidth]{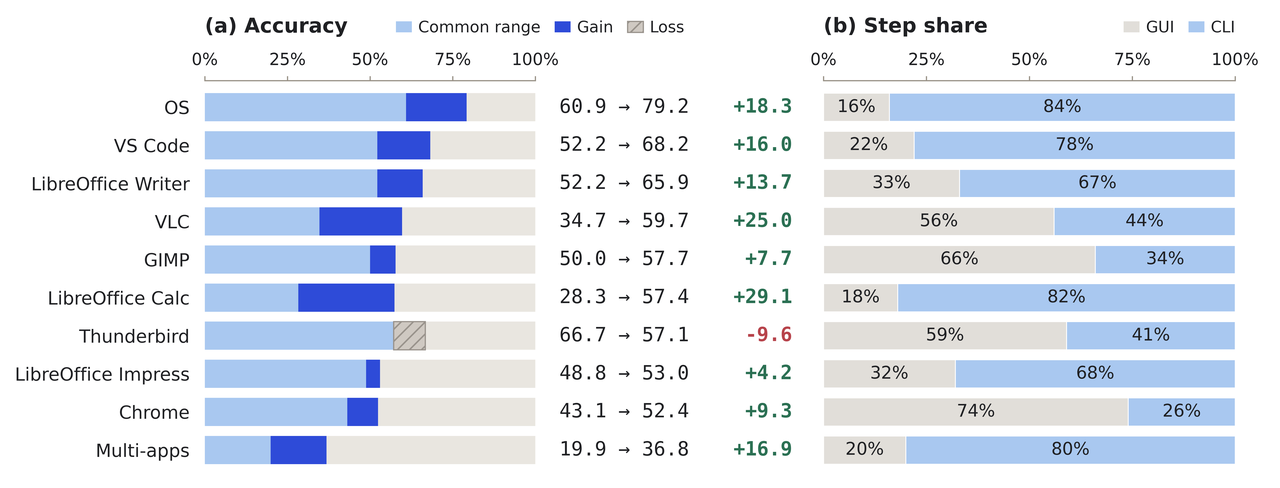}
    \caption{Domain-level accuracy and GUI/CLI step shares on OSWorld.}
    \label{fig:rq1_interface_usage}
\end{figure}

\begin{figure}[!htb]
    \centering
    \includegraphics[width=\linewidth]{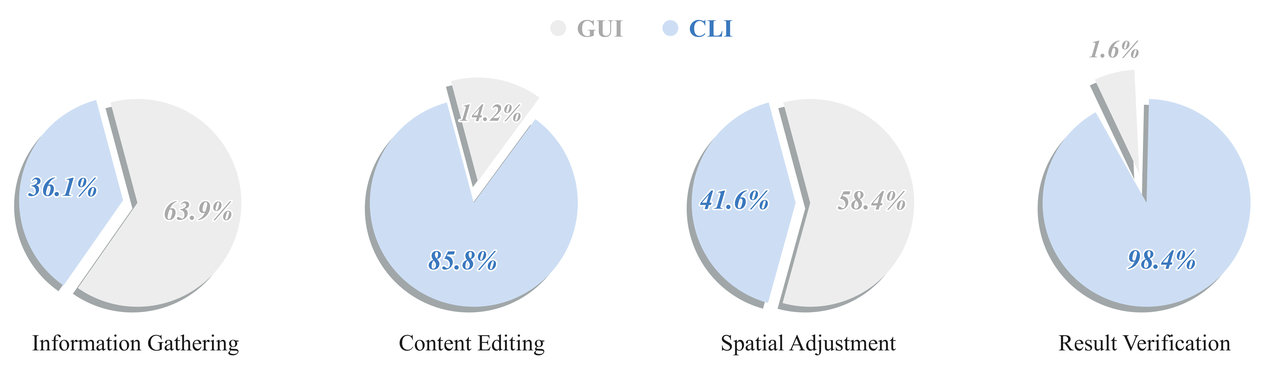}
    \caption{GUI/CLI step shares by operation category for HybridCUA-9B.}
    \label{fig:rq1_operation_purposes}
\end{figure}

\paragraph{Interface selection.}
\label{sec:interface_selection}
Figure~\ref{fig:rq1_interface_usage} suggests that HybridCUA selects
interfaces according to domain specific interaction needs: CLI dominates
OS tasks (84\%), where shell commands provide direct access to
files and system settings, whereas GUI dominates Chrome tasks (74\%),
where interaction centers on web pages.
The breakdown in Figure~\ref{fig:rq1_operation_purposes}
links these preferences to operation requirements: CLI dominates content
editing (85.8\%) and result verification (98.4\%), which suit direct data
manipulation and state inspection, while GUI is favored for spatial
adjustment (58.4\%), where visual feedback guides positioning and layout.

\begin{figure}[!tbp]
    \centering
    \includegraphics[width=\linewidth]{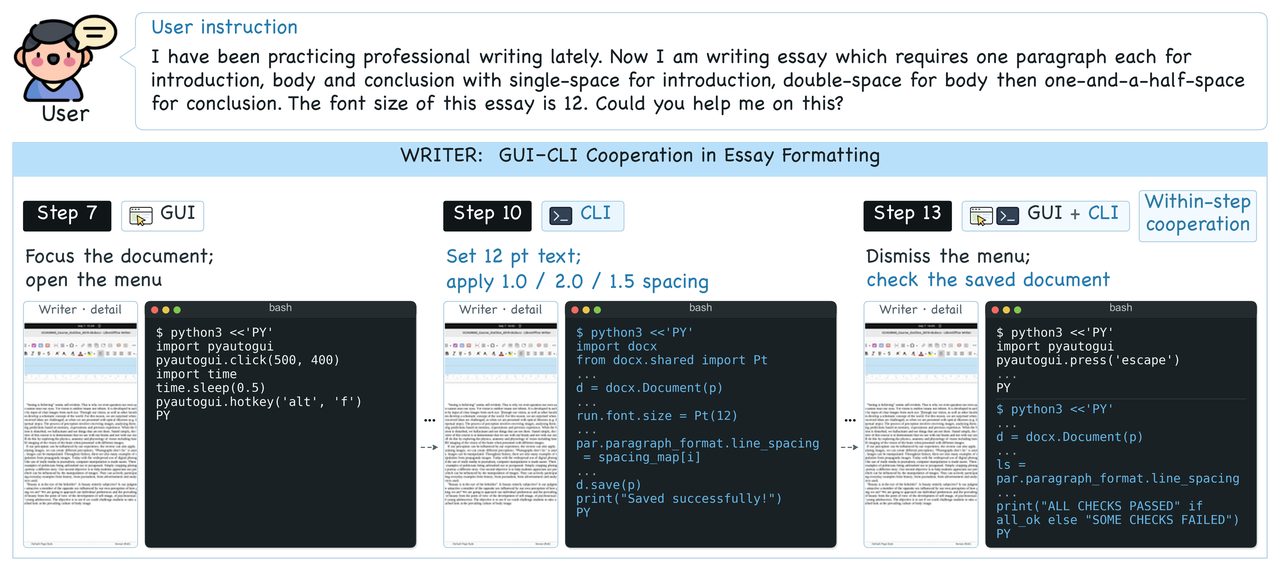}
    \caption{GUI--CLI cooperation in essay formatting. Code is excerpted
    and screenshots are cropped.}
    \label{fig:interface_cooperation}
\end{figure}

\paragraph{Complementary GUI--CLI cooperation.}
\label{sec:interface_cooperation}
Figure~\ref{fig:interface_cooperation} illustrates \textbf{two levels of
GUI--CLI cooperation} in an essay-formatting task. \textbf{Across the
trajectory}, GUI steps focus the Writer window and open its menu
(step~7), while CLI steps use python-docx to apply 12 point text and
single, double, and one-and-a-half line spacing to the introduction,
body, and conclusion, respectively, before saving the file (step~10).
These steps combine application interaction with direct document editing
to advance the same task. \textbf{Within a hybrid step}, step~13 first
dismisses the menu using pyautogui, then reopens the saved file through
the CLI to check font size, line spacing, and spacing rules. Both
operations execute sequentially in a single bash call, combining GUI
state handling with file level verification within one step.

\setlength{\columnsep}{8pt}
\begin{wraptable}{r}{0.48\linewidth}
    \centering
    \small
    \caption{Effect of action schema on OSWorld.}
    \label{tab:action_schema_analysis}
    \setlength{\tabcolsep}{4pt}
    \renewcommand{\arraystretch}{1.05}
    \begin{tabular}{@{}lcc@{}}
        \hline
        Configuration & Acc.$\uparrow$ & Avg. Steps$\downarrow$ \\
        \hline
        Qwen3.5-9B & 38.8 & 31.6 \\
        Separate GUI/CLI tools & 38.8 & 16.8 \\
        \rowcolor[RGB]{220,232,247}
        \textbf{Unified \texttt{bash} action (ours)} & \textbf{46.0} & \textbf{19.8} \\
        \hline
    \end{tabular}
\end{wraptable}

\paragraph{Unified action schema matters.}
\label{sec:action_schema_analysis}
We isolate the effect of the action schema by fine-tuning two variants
from the same Qwen3.5-9B base and trajectories under matched settings. The
separate-tool schema keeps the two interfaces in distinct tools,
\texttt{computer\_use} for GUI actions and \texttt{cli} for shell actions,
whereas the unified schema expresses both through a single \texttt{bash}
action.
As shown in Table~\ref{tab:action_schema_analysis}, the unified format achieves
\textbf{46.0\%} accuracy with \textbf{19.8} average steps, outperforming the
separate-tool schema by \textbf{7.2} percentage points at the cost of
\textbf{3.0} additional steps. It also improves over the GUI only base, indicating that a shared action grammar improves task
completion while remaining more efficient than the base.

\section{Conclusion}
\label{sec:conclusion}

We introduced \textbf{HybridCUA}, a framework that teaches computer-use agents
\emph{when} and \emph{how} to combine GUI and CLI actions. HybridCUA constructs
three types of trajectories together with verified RLVR tasks, and trains a unified policy model through
supervised fine-tuning and
CLI-aware online reinforcement learning. Experiments across multiple computer-use
benchmarks show that HybridCUA improves both task completion and interaction
efficiency over its base model, while generalizing across interfaces and
operating systems. These results demonstrate that the GUI and CLI provide
complementary capabilities and that learning to orchestrate them offers a
promising direction for accurate, efficient, and generalizable computer-use
agents.

\section*{AI Use Statement}

We used AI assistants in a limited supporting role for language polishing,
LATEX formatting, coding, and debugging. Large language models are also
part of our methodology, as our data construction pipeline samples CLI only
and interleaved trajectories from strong models through an execution harness
(Section~\ref{sec:trajectory_construction}). The authors reviewed AI-assisted
outputs before incorporating them into the paper. The authors made the final
decisions regarding the methodology, experiments, analyses, and presentation,
and take full responsibility for the content and results of this work.

\section*{Ethics Statement}

This work studies hybrid GUI and CLI orchestration for computer-use agents in
benchmark computer environments. The experiments do not involve human
participants or the collection of private or personally identifiable
information. All trajectory collection, training, and evaluation are carried
out in sandboxed virtual machines that are reset between episodes and have no
access to real user accounts or credentials. Because shell access amplifies
what an agent can do in a single action, behavior and safety in real-world
deployments are beyond the scope of this study; such deployments would require
further evaluation, safeguards, and human oversight.

\section*{Reproducibility Statement}

Section~\ref{sec:method} describes the hybrid formulation, the data generation
pipeline, and the two-stage training paradigm with the CLI-aware reward.
Appendix~\ref{app:action_space} specifies the action space and interaction
protocol, and Appendix~\ref{app:dataset_details} details trajectory
construction, RLVR task labeling, and dataset statistics.
Appendix~\ref{app:training_details} reports the supervised and RL
hyperparameters and compute setup, while
Appendix~\ref{app:evaluation_protocol} documents the benchmark settings,
baselines, and metric definitions, and Appendix~\ref{app:evaluation_prompt}
provides the system prompts used for evaluation and data construction. We will
release the trajectories, RLVR tasks, data generation and training pipelines,
and the HybridCUA-9B model.

\bibliography{iclr2027_conference}

\begin{thebibliography}{33}
\providecommand{\natexlab}[1]{#1}
\providecommand{\url}[1]{\texttt{#1}}
\expandafter\ifx\csname urlstyle\endcsname\relax
  \providecommand{\doi}[1]{doi: #1}\else
  \providecommand{\doi}{doi: \begingroup \urlstyle{rm}\Url}\fi

\bibitem[Bai et~al.(2025)Bai, Cai, Chen, et~al.]{bai2025qwen3vl}
Shuai Bai, Yuxuan Cai, Ruizhe Chen, et~al.
\newblock {Qwen3-VL} technical report.
\newblock \emph{arXiv preprint arXiv:2511.21631}, 2025.

\bibitem[Bai et~al.(2026)Bai, Deng, Fan, et~al.]{bai2026recreationworld}
Shuai Bai, Jiayong Deng, Sicheng Fan, et~al.
\newblock {RecreationWorld}: Scalable and verifiable environments for hybrid computer-use agents.
\newblock \emph{arXiv preprint arXiv:2609.22000}, 2026.

\bibitem[Bonatti et~al.(2024)Bonatti, Zhao, Bonacci, et~al.]{bonatti2024windowsagentarena}
Rogerio Bonatti, Dan Zhao, Francesco Bonacci, et~al.
\newblock Windows agent arena: Evaluating multi-modal os agents at scale.
\newblock \emph{arXiv preprint arXiv:2409.08264}, 2024.

\bibitem[{Claude Code Team}(2026)]{claudecode2026}
{Claude Code Team}.
\newblock {Claude Code}.
\newblock GitHub repository, 2026.
\newblock URL \url{https://github.com/anthropics/claude-code}.

\bibitem[Hu et~al.(2026)Hu, Zhang, Xu, et~al.]{hu2026toolcua}
Xuhao Hu, Xi~Zhang, Haiyang Xu, et~al.
\newblock {ToolCUA}: Towards optimal gui-tool path orchestration for computer use agents.
\newblock \emph{arXiv preprint arXiv:2605.12481}, 2026.

\bibitem[Jia et~al.(2025)Jia, Liao, Zhang, Xu, Xie, Jiang, Yan, Liu, Ye, and Huang]{jia2025osworldmcp}
Hongrui Jia, Jitong Liao, Xi~Zhang, Haiyang Xu, Tianbao Xie, Chaoya Jiang, Ming Yan, Si~Liu, Wei Ye, and Fei Huang.
\newblock {OSWorld-MCP}: Benchmarking mcp tool invocation in computer-use agents.
\newblock \emph{arXiv preprint arXiv:2510.24563}, 2025.

\bibitem[Lai et~al.(2025)Lai, Liu, Zhao, et~al.]{lai2025computerrl}
Hanyu Lai, Xiao Liu, Yanxiao Zhao, et~al.
\newblock {ComputerRL}: Scaling end-to-end online reinforcement learning for computer use agents.
\newblock \emph{arXiv preprint arXiv:2508.14040}, 2025.

\bibitem[Li et~al.(2026)Li, Zhou, Yu, Xu, Yang, Li, and Shan]{li2026weavebench}
Wanli Li, Bowen Zhou, Yunyao Yu, Zhou Xu, Yifan Yang, Dongsheng Li, and Caihua Shan.
\newblock {WeaveBench}: A long-horizon, real-world benchmark for computer-use agents with hybrid interfaces.
\newblock \emph{arXiv preprint arXiv:2606.09426}, 2026.

\bibitem[Lian et~al.(2026)Lian, Chen, Yu, Duan, Liu, Liu, Fu, Luan, Qu, Xia, and Wang]{lian2026uimopd}
Niu Lian, Tongbo Chen, Zhehao Yu, Chengzhen Duan, Fazhan Liu, Hui Liu, Pei Fu, Jian Luan, Heng Qu, Shu-Tao Xia, and Jinpeng Wang.
\newblock {UI-MOPD}: Multi-platform on-policy distillation for unified gui agents.
\newblock \emph{arXiv preprint arXiv:2607.04425}, 2026.

\bibitem[Lu et~al.(2025)Lu, Ye, Tang, Shen, Xu, Zheng, Lu, Yan, Huang, Xiao, et~al.]{lu2025uis1}
Zhengxi Lu, Jiabo Ye, Fei Tang, Yongliang Shen, Haiyang Xu, Ziwei Zheng, Weiming Lu, Ming Yan, Fei Huang, Jun Xiao, et~al.
\newblock Ui-s1: Advancing gui automation via semi-online reinforcement learning.
\newblock \emph{arXiv preprint arXiv:2509.11543}, 2025.

\bibitem[Lu et~al.(2026{\natexlab{a}})Lu, Chai, Guo, Yin, Liu, Wang, Xiao, Ren, Zhao, Liu, et~al.]{lu2026uir1}
Zhengxi Lu, Yuxiang Chai, Yaxuan Guo, Xi~Yin, Liang Liu, Hao Wang, Han Xiao, Shuai Ren, Pengxiang Zhao, Guangyi Liu, et~al.
\newblock Ui-r1: Enhancing efficient action prediction of gui agents by reinforcement learning.
\newblock In \emph{Proceedings of the AAAI Conference on Artificial Intelligence}, volume~40, pp.\  17608--17616, 2026{\natexlab{a}}.

\bibitem[Lu et~al.(2026{\natexlab{b}})Lu, Tang, Liu, Ma, Song, Tan, Zhang, Lu, Xiao, Zhuang, et~al.]{lu2026uicopilot}
Zhengxi Lu, Fei Tang, Guangyi Liu, Jin Ma, Kaitao Song, Xu~Tan, Wenqi Zhang, Weiming Lu, Jun Xiao, Yueting Zhuang, et~al.
\newblock Ui-copilot: Advancing long-horizon gui automation via tool-integrated policy optimization.
\newblock In \emph{Proceedings of the 64th Annual Meeting of the Association for Computational Linguistics (Volume 1: Long Papers)}, pp.\  19741--19762, 2026{\natexlab{b}}.

\bibitem[Lv et~al.(2026)Lv, Liu, Ren, Lai, Jing, Zhang, Zhao, Yao, Tang, and Dong]{lv2026scalecua}
Bowen Lv, Xiao Liu, Yanyu Ren, Hanyu Lai, Bohao Jing, Hanchen Zhang, Yanxiao Zhao, Shuntian Yao, Jie Tang, and Yuxiao Dong.
\newblock {SCALECUA}: Scaling computer use agents with verifiable task synthesis and efficient online rl.
\newblock \emph{arXiv preprint arXiv:2607.11185}, 2026.

\bibitem[{OpenAI}(2026)]{openai2026codex}
{OpenAI}.
\newblock {Codex for (almost) everything}.
\newblock OpenAI, April 2026.
\newblock URL \url{https://openai.com/index/codex-for-almost-everything/}.
\newblock Published April 16, 2026.

\bibitem[{OpenClaw Contributors}(2026)]{openclaw2026peekaboo}
{OpenClaw Contributors}.
\newblock {Peekaboo}: Mac automation that sees the screen and does the clicks.
\newblock GitHub repository, 2026.
\newblock URL \url{https://github.com/openclaw/Peekaboo}.

\bibitem[Qin et~al.(2025)Qin, Ye, Fang, et~al.]{qin2025uitars}
Yujia Qin, Yining Ye, Junjie Fang, et~al.
\newblock {UI-TARS}: Pioneering automated gui interaction with native agents.
\newblock \emph{arXiv preprint arXiv:2501.12326}, 2025.

\bibitem[{Qwen Team}(2026{\natexlab{a}})]{qwen2026qwen35}
{Qwen Team}.
\newblock {Qwen3.5}: Towards native multimodal agents.
\newblock Qwen Blog, 2026{\natexlab{a}}.
\newblock URL \url{https://qwen.ai/blog?id=qwen3.5}.
\newblock Published February 15, 2026.

\bibitem[{Qwen Team}(2026{\natexlab{b}})]{qwen2026qwen38}
{Qwen Team}.
\newblock {Qwen3.8}.
\newblock Qwen Blog, 2026{\natexlab{b}}.
\newblock URL \url{https://qwen.ai/blog?id=qwen3.8}.
\newblock {Qwen3.8-27B} open-weight release, August 14, 2026.

\bibitem[Sheng et~al.(2025)Sheng, Zhang, Ye, Wu, Zhang, Zhang, Peng, Lin, and Wu]{sheng2025hybridflow}
Guangming Sheng, Chi Zhang, Zilingfeng Ye, Xibin Wu, Wang Zhang, Ru~Zhang, Yanghua Peng, Haibin Lin, and Chuan Wu.
\newblock {HybridFlow}: A flexible and efficient {RLHF} framework.
\newblock In \emph{Proceedings of the Twentieth European Conference on Computer Systems (EuroSys)}, 2025.

\bibitem[Shi et~al.(2026)Shi, Wang, Fang, Liang, Jin, Zhao, Liu, Chen, and Wang]{shi2026cuauniverse}
Haoting Shi, Wenhao Wang, Weicheng Fang, Yaozhong Liang, Tian Jin, Pengxiang Zhao, Guangyi Liu, Siheng Chen, and Yanfeng Wang.
\newblock {CUA-Universe}: A scalable and dynamic environment for hybrid {GUI+CLI} agents.
\newblock \emph{arXiv preprint arXiv:2609.05374}, 2026.

\bibitem[Shoeybi et~al.(2019)Shoeybi, Patwary, Puri, LeGresley, Casper, and Catanzaro]{shoeybi2019megatron}
Mohammad Shoeybi, Mostofa Patwary, Raul Puri, Patrick LeGresley, Jared Casper, and Bryan Catanzaro.
\newblock {Megatron-LM}: Training multi-billion parameter language models using model parallelism.
\newblock \emph{arXiv preprint arXiv:1909.08053}, 2019.

\bibitem[{THUDM}(2025)]{thudm2025slime}
{THUDM}.
\newblock slime: An {LLM} post-training framework for {RL} scaling.
\newblock \url{https://github.com/THUDM/slime}, 2025.

\bibitem[Wang et~al.(2026)Wang, Lu, Wang, Bai, Liu, Zhang, Wang, Hu, Xie, Bai, Liu, Shen, Lin, and Yu]{wang2026cuagym}
Bowen Wang, Dunjie Lu, Junli Wang, Tianyi Bai, Shixuan Liu, Zhipeng Zhang, Haiquan Wang, Hao Hu, Tianbao Xie, Shuai Bai, Dayiheng Liu, Que Shen, Junyang Lin, and Tao Yu.
\newblock {CUA-Gym}: Scaling verifiable training environments and tasks for computer-use agents.
\newblock \emph{arXiv preprint arXiv:2605.25624}, 2026.

\bibitem[Wang et~al.(2025{\natexlab{a}})Wang, Zou, Song, et~al.]{wang2025uitars2}
Haoming Wang, Haoyang Zou, Huatong Song, et~al.
\newblock {UI-TARS-2} technical report: Advancing gui agent with multi-turn reinforcement learning.
\newblock \emph{arXiv preprint arXiv:2509.02544}, 2025{\natexlab{a}}.

\bibitem[Wang et~al.(2025{\natexlab{b}})Wang, Wang, Lu, et~al.]{wang2025opencua}
Xinyuan Wang, Bowen Wang, Dunjie Lu, et~al.
\newblock {OpenCUA}: Open foundations for computer-use agents.
\newblock \emph{arXiv preprint arXiv:2508.09123}, 2025{\natexlab{b}}.

\bibitem[Wu et~al.(2024)Wu, Wu, Xu, et~al.]{wu2024osatlas}
Zhiyong Wu, Zhenyu Wu, Fangzhi Xu, et~al.
\newblock {OS-ATLAS}: A foundation action model for generalist gui agents.
\newblock \emph{arXiv preprint arXiv:2410.23218}, 2024.

\bibitem[Xie et~al.(2024)Xie, Zhang, Chen, et~al.]{xie2024osworld}
Tianbao Xie, Danyang Zhang, Jixuan Chen, et~al.
\newblock {OSWorld}: Benchmarking multimodal agents for open-ended tasks in real computer environments.
\newblock \emph{arXiv preprint arXiv:2404.07972}, 2024.

\bibitem[Xue et~al.(2026)Xue, Peng, Huang, Guo, Han, Wang, Wang, Zhang, Yang, Zhao, Ding, Ma, Xie, Pei, Cai, and Qiu]{xue2026evocua}
Taofeng Xue, Chong Peng, Mianqiu Huang, Linsen Guo, Tiancheng Han, Haozhe Wang, Jianing Wang, Xiaocheng Zhang, Xin Yang, Dengchang Zhao, Jinrui Ding, Xiandi Ma, Yuchen Xie, Peng Pei, Xunliang Cai, and Xipeng Qiu.
\newblock {EvoCUA}: Evolving computer use agents via learning from scalable synthetic experience.
\newblock \emph{arXiv preprint arXiv:2601.15876}, 2026.

\bibitem[Yan et~al.(2025)Yan, Wang, Du, et~al.]{yan2025mcpworld}
Yunhe Yan, Shihe Wang, Jiajun Du, et~al.
\newblock {MCPWorld}: A unified benchmarking testbed for api, gui, and hybrid computer use agents.
\newblock \emph{arXiv preprint arXiv:2506.07672}, 2025.

\bibitem[Yang et~al.(2025)Yang, Yang, Dou, et~al.]{yang2025ultracua}
Yuhao Yang, Zhen Yang, Zi-Yi Dou, et~al.
\newblock {UltraCUA}: A foundation model for computer use agents with hybrid action.
\newblock \emph{arXiv preprint arXiv:2510.17790}, 2025.

\bibitem[Yang et~al.(2026)Yang, Fan, and Huang]{yang2026clianything}
Yuhao Yang, Tianyu Fan, and Chao Huang.
\newblock {CLI-Anything}: Towards agent-native computer use.
\newblock \emph{arXiv preprint arXiv:2606.03854}, 2026.

\bibitem[Zheng et~al.(2024)Zheng, Yin, Xie, Sun, Huang, Yu, Cao, Kozyrakis, Stoica, Gonzalez, Barrett, and Sheng]{zheng2024sglang}
Lianmin Zheng, Liangsheng Yin, Zhiqiang Xie, Chuyue Sun, Jeff Huang, Cody~Hao Yu, Shiyi Cao, Christos Kozyrakis, Ion Stoica, Joseph~E. Gonzalez, Clark Barrett, and Ying Sheng.
\newblock {SGLang}: Efficient execution of structured language model programs.
\newblock In \emph{Advances in Neural Information Processing Systems (NeurIPS)}, 2024.

\bibitem[Zhou et~al.(2026)Zhou, Tong, Zhang, Kong, Cai, Xia, Zhang, Zhang, Li, Chen, Wang, Dai, Li, Chen, Wang, and Hoi]{zhou2026qwen_ui_agent}
Hanzhang Zhou, Panrong Tong, Xu~Zhang, Quyu Kong, Chenglin Cai, Tianyu Xia, Gongjie Zhang, Jianan Zhang, Long Li, Long Chen, Lei Wang, Gaole Dai, Pengxiang Li, Liangyu Chen, Yue Wang, and Steven Hoi.
\newblock {Qwen-UI-Agent} technical report: Toward next-generation real-world centric foundation {GUI} agents.
\newblock \emph{arXiv preprint arXiv:2607.28227}, 2026.

\end{thebibliography}
\bibliographystyle{iclr2027_conference}

\clearpage
\appendix

\end{document}